\documentclass[10pt, a4paper]{article}

\usepackage{lrec2026}

\usepackage{subcaption}
\usepackage{amsfonts} 
\usepackage{amssymb} 
\usepackage{amsmath} 
\usepackage{enumitem}
\usepackage{listings}

\usepackage{tabularx}
\usepackage{booktabs}
\usepackage[utf8]{inputenc}
\usepackage{microtype}
\usepackage{inconsolata}
\usepackage{graphicx}
\usepackage{multirow}
\usepackage{float}

\usepackage{annotates}
\providecommentcommand{rom}{blue}{Roman}
\providecommentcommand{yar}{red}{Yarik}

\definecolor{customred}{RGB}{255, 0, 100}
\definecolor{customgreen}{RGB}{0, 128, 0}

\title{PARL: \underline{P}rompt-based \underline{A}gents for \underline{R}einforcement \underline{L}earning}

\name{Yarik Menchaca Resendiz$^{1,2}$, Roman Klinger$^{2}$}

\address{$^{1}$Leibniz-Institut f\"ur Psychologie (ZPID), Trier, Germany\\
  $^{2}$Fundamentals of Natural Language Processing, University of Bamberg, Germany\\
  ymr@leibniz-psychology.org, roman.klinger@uni-bamberg.de}

\abstract{%
  Large language models (LLMs) have demonstrated high performance on tasks expressed in natural language, particularly in zero- or few-shot settings. These are typically framed as supervised (e.g., classification) or unsupervised (e.g., clustering) problems. However, limited work evaluates LLMs as agents in reinforcement learning (RL) tasks (e.g., playing games), where learning occurs through interaction with an environment and a reward system. While prior work focused on representing tasks that rely on a language representation, we study structured, non-linguistic reasoning -- such as interpreting positions in a grid world. We therefore introduce PARL (Prompt-based Agent for Reinforcement Learning), a method that uses LLMs as RL agents through prompting, without any fine-tuning. PARL encodes actions, states, and rewards in the prompt, enabling the model to learn through trial-and-error interaction. We evaluate PARL on three standard RL tasks that do not entirely rely on natural language. We show that it can match or outperform traditional RL agents in simple environments by leveraging pretrained knowledge. However, we identify performance limitations in tasks that require complex mathematical operations or decoding states and actions.
  \\ \newline%
}

\begin{document}

\maketitleabstract

\section{Introduction}
\label{sec:intro}
Large language models~\citep[LLMs; e.g.,][]{openai2024gpt4, jiang2024mixtral, devlin-etal-2019-bert} have demonstrated strong performance across a wide range of tasks. While originally developed for text-based applications, they are being applied to other modalities such as vision~\citep{dosovitskiy2021an, pmlr-v139-radford21a} and audio~\citep{Borsos2022AudioLMAL, rubenstein2023audiopalm}. In natural language processing (NLP), tasks are often framed as supervised learning problems, relying on labeled data, such as classification~\citep{sun-etal-2023-text, schick-schutze-2021-exploiting} and translation~\citep{zhang-etal-2023-machine, moslem-etal-2023-adaptive, brants-etal-2007-large}, or as unsupervised learning problems, such as text clustering~\citep{zhang-etal-2023-clusterllm, viswanathan-etal-2024-large}. Although LLMs have addressed many such tasks using (zero- and few-shot) prompting, they are predominantly based on supervised or unsupervised learning frameworks.

\begin{figure}[t]
  \centering
  \includegraphics[width=.8\columnwidth]{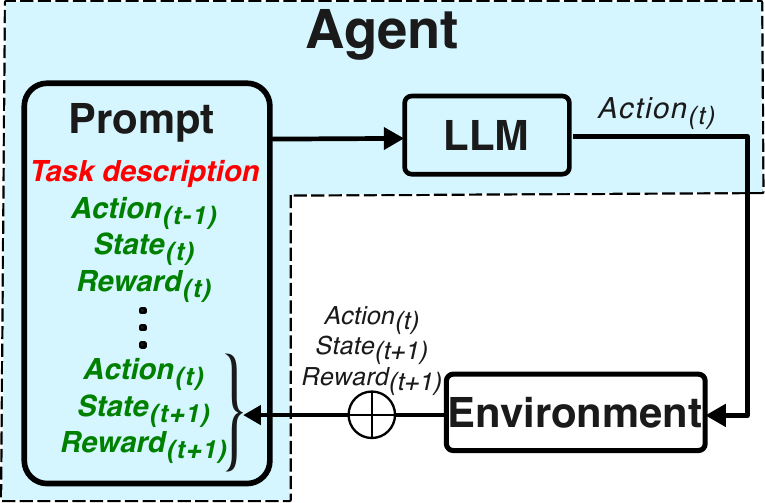}
  \caption{Prompt-based Agent Optimization for Reinforcement Learning (PARL)
    uses a large language model to make decisions through in-context learning. The optimization begins with a prompt that contains only the
    \textbf{\textcolor{customred}{task description}}. As the agent
    interacts with the environment, the prompt is updated by concatenating ($\oplus$) the history of interactions of the
    \textbf{\textcolor{customgreen}{states, actions, and rewards}}
    generated at each step $t$.}
  \label{fig:depiction}
\end{figure}

Many real-world problems are, however, more appropriately modeled as reinforcement learning~\citep[RL;][]{kreutzer-etal-2021-offline}. In RL, agents learn by interacting with an environment, and take actions and receive feedback in the form of rewards. Unlike supervised learning, which relies on labeled data, RL involves trial-and-error learning with delayed rewards and requires balancing exploration and exploitation. Common applications include learning to play games (e.g., Go~\citep{silver2016mastering}, Atari games~\citep{schrittwieser2020mastering}, Chess~\citep{david2016deepchess}, BlackJack~\citep{zha2019rlcard}), robotics~\citep{kober2013reinforcement}), and improving large models with human feedback~\citep{christiano2017deep, kaufmann2023survey}.

While recent work has explored LLM-based agents for decision-making via techniques like chain-of-thought prompting or self-reflection~\citep{yao2023react, zhao2023large, NEURIPS2023_b5c8c1c1, shinn2023reflexion}, these methods treat LLMs as text-based agents that rely on natural language for both input and output. Such designs are effective for language-centered tasks (e.g., planning), but their applicability to structured, non-linguistic environments remains underexplored.

In this paper, we investigate and evaluate whether LLMs can act as traditional RL agents in tasks where input and output are not naturally expressed in natural language -- such as navigating a grid world. We introduce PARL (\underline{P}rompt-based \underline{A}gent for \underline{R}einforcement \underline{L}earning), a method that keeps LLM weights frozen and uses prompting to enable in-context learning through iterative interaction with an environment. At each step, PARL encodes states, actions, and rewards into a cumulative text prompt, allowing the model to improve decision-making via in-context learning (Figure~\ref{fig:depiction}).

We evaluate PARL against three state-of-the-art RL policies across three established RL tasks (Blackjack, Frozen Lake, and Taxi) to address: (RQ1) \textit{Can a PARL agent learn from interactions with the environment similarly to other RL policies?}: (RQ2) \textit{Can a PARL agent benefit from pre-trained knowledge from the LLMs?}; (RQ3) \textit{ Does a PARL agent explore and exploit similarly to a standard RL agent?} Our findings show that PARL can match or outperform established policies in knowledge-intensive tasks (e.g., blackjack), where the agent benefits from the LLM’s pre-trained information. However, in more complex tasks, PARL faces challenges due to LLMs' limitations in symbolic computation and state decoding.

\section{Related Work}
In Section~\ref{subsec:related_work_rl}, we review state-of-the-art reinforcement learning methods, whereas in Section~\ref{subsec:llms-in-rl}, we discuss the use of LLMs as agents in RL.

\subsection{Reinforcement Learning}
\label{subsec:related_work_rl}

Reinforcement learning aims at optimizing agents' actions in sequential decision-making tasks through trial-and-error learning guided by a reward system.
The problem is typically formalized as a Markov decision process, defined by $( S, A, P, R, \gamma)$, where $S$ represents the state space, $A$ the action space, $P(s'|s, a)$ the state transition probability, $R(s, a)$ the reward function, and $\gamma \in [0, 1]$ the discount factor.
The goal is to find an optimal policy $\pi(a|s)$ that maximizes the expected cumulative return:
\begin{equation}
  G = \sum_{t=0}^\infty \gamma^t R(s_t, a_t)
\end{equation}
Q-learning and Deep Q-Networks~\citep[DQN;][]{Mnih2013PlayingAW, mnih2015human} set foundational work by integrating neural networks to approximate the action-value function $Q(s, a)$, which represents the expected reward for taking action $a$ in state $s$.
In DQN, this function is parameterized as $Q(s, a; \theta)$, where $\theta$ denotes the weights of the neural network.
The model learns these parameters by minimizing a sequence of loss functions $\mathcal{L}_i(\theta_i)$ that changes at each iteration $i$:
\begin{equation}
  \mathcal{L}^{\text{DQN}}(\theta) = \mathbb{E}_{s, a \sim p(\cdot)}
  \left[\left(y_i - Q(s, a; \theta)\right)^2\right],
\end{equation}
where {\small$y_i~=~\mathbb{E}_{s' \sim \mathcal{E}}
  \left[r+\gamma\max_{a'} Q(s', a'; \theta_{i-1})\mid s, a\right]$}
is the target for iteration $i$ and $p(s, a)$
is a probability distribution over sequences $s$ and actions $a$.

Proximal Policy Optimization \citep[PPO;][]{schulman2017proximal}
balances exploration and exploitation while maintaining a stable policy
updates by taking small policy updates -- big updates may guide the policy
in a suboptimal direction. PPO uses a clipped objective function to ensure
small policy updates, defined as:
\begin{multline}
  \mathcal{L}^{\text{PPO}}(\theta) = \mathbb{E}_t\Big[
  \min \Big(
  r_t(\theta) \hat{A}_t, \\
  \text{clip}(r_t(\theta), 1 \pm \epsilon) \hat{A}_t
  \Big)
  \Big],
\end{multline}
where $r_t(\theta)$ is the probability ratio, $\hat{A}_t$
the advantage estimate, and $\epsilon$ the clipping parameter.

Advantage Actor-Critic \citep[A2C;][]{pmlr-v48-mniha16}
combines value- and policy-based methods in an actor-critic framework, where
the actor $\pi_\theta(a \mid s)$ selects actions and the critic $V(s)$
estimates the value function to reduce variance, improving learning
stability. The A2C loss function is defined as:
\begin{equation}
  \mathcal{L}^{\text{A2C}}(\theta) = \mathbb{E}_t
  \left[ \log \pi_\theta(a_t|s_t) \hat{A}_t \right],
\end{equation}
where $\hat{A}_t = r_t + \gamma V(S_{t+1}) - V(s)$
is the advantage function computed from the critic's value estimated.

\subsection{Prompting in Reinforcement Learning}
\label{subsec:llms-in-rl}

LLMs, such as GPT-4~\citep{openai2024gpt4}, LLaMa 3~\citep{grattafiori2024llama3herdmodels}, and Mixtral~\citep{jiang2024mixtral}, have demonstrated strong performance on zero- and few-shot tasks using prompts~\citep{semnani-etal-2023-wikichat, lin-etal-2022-shot}. These prompting strategies are widely used in traditional NLP tasks. For example, text classification prompts combine instructions with labels~\citep[e.g.,\textit{``Tag the text as positive or negative\ldots''};][] {hu-etal-2022-knowledgeable, gu-etal-2022-ppt}, while summarization include keywords like \textit{``TL; DR''} or \textit{``summarize''}~\citep{radford2019language, narayan-etal-2021-planning}. Translation prompts specify source and target languages~\citep[e.g., \textit{``Translate English to German''};][]{raffel2019exploring}.

While these prompting strategies have primarily been applied to traditional
(un)supervised tasks, recent work has explored how LLMs can be integrated into RL as support decision-making algorithms. For instance, \citet{zhao2023large} propose LLM-MCTS, an algorithm that uses LLMs as a
commonsense-informed world model and a heuristic policy within the Monte Carlo Tree Search (MCTS) framework. Their approach outperforms
standard MCTS and purely LLM-based policies in complex planning tasks. Similarly, AdaPlanner~\citep{NEURIPS2023_b5c8c1c1} proposes a method that incrementally refines plans -- represented as prompts containing a sequence of high-level steps (e.g., clean lettuce on the dining table) -- based on feedback from the environment. This feedback is evaluated by comparing predicted outcomes against actual outcomes, allowing the model to adjust its plan when misalignment occurs.

Other works aim to improve the LLM-based agents through reasoning (e.g., chain-of-thought) and learning from experience~\citep{zhang-etal-2024-agent}. ReAct \citep{yao2023react} interleaves reasoning with actions (interactions with tools or environments), which improves task performance and interpretability while reducing hallucinations. Reflexion \citep{shinn2023reflexion} extends this by enabling agents to self-critique (e.g.,  critique their own decisions) and update their behavior over time through memory-based adaptation, improving performance without additional fine-tuning. This self-improving behavior is effective in domains like ALFWorld (an environment for instruction-following tasks) and HotPotQA (question answering across documents).

To evaluate LLMs' capabilities, recent benchmarks use structured, game-like environments. Clembench~\citep{chalamalasetti-etal-2023-clembench} exposes models to constrained decision-making tasks to test instruction-following and consistency. Results indicate that chat-optimized LLMs struggle with simple interactive scenarios, revealing limitations in current agent designs. Similarly, \citet{electronics13081532} use games like Tic-Tac-Toe for evaluation, benefiting from well-defined rules and outcomes.

Despite this progress, most prior work uses LLMs as text-based agents, where both input and output are in natural language. While this is effective for tasks like planning or instruction following, it has not been explored or evaluated in settings with structured, non-linguistic inputs -- such as grid positions or card values in games like blackjack.

\section{Methods}
In the following section, we introduce PARL (in Section~\ref{subsection:
  PromptRL-Agent}).  PARL adopts the standard RL framework, learning
through interaction with the environment to achieve task-specific
goals (in Section~\ref{subsection: Training})\footnote{Code and resources
  available at \url{https://github.com/YarikMR/PARL}}.

\subsection{PARL: Prompt Base Agent for Reinforcement Learning}
\label{subsection: PromptRL-Agent}
PARL represents a reinforcement learning policy optimization method
that uses LLMs as decision-making agents via prompting techniques. It
combines the task description $\mathcal{T}$
with the concatenated interaction history ($h$) with the environment. The PARL policy is defined by the prompt:
\begin{equation}
  \label{equation:pal}
  \mathcal{P}^{\text{PARL}} = \mathcal{T} \bigoplus_{t=0}^n h_t,
\end{equation}
where $\bigoplus$ indicates concatenation and $\mathcal{T}$ -- the \textit{task description} -- is defined as:
\[
  \mathcal{T} = (\mathcal{G}, \mathcal{A}, \mathcal{S}, \mathcal{R}),
\]
where $\mathcal{G}$ denotes the task goal. For example, in the context of Blackjack: \textit{``Blackjack is a card game where the goal is to get as close to 21 as
  possible without exceeding it.''}. \(\mathcal{A}\) represents the set of actions available \(\{a_1, \dots, a_k\}\) to the agent (e.g., \textit{``0: Stick (Stop), 1: Hit (Draw)''}). \(\mathcal{S}\) specifies the state representation, describing the information observed by the agent (e.g., \textit{``The observation is a tuple: (player’s value, dealer’s value)''}).
Finally, \(\mathcal{R}\) is the set of rewards \(\{r_1, \dots, r_k\}\) (e.g., \textit{``+1: Win, --1: Lose, 0: Draw.''}). $h_t$ represents the history of interactions with the environment, defined as:
\[
  h_t = (\text{s}_t, \text{a}_t, \text{r}_t),
\]
where \(\text{s}_t\) denotes the state at time \(t\), \(\text{a}_t\)
represents the action taken at time \(t\), and \(\text{r}_t\) is the reward received as a result of the action. This iterative log captures the sequence of interactions, recording the state, action, and reward for each time step \(t = 0, \dots, n\). For example: \textit{``Action (taken): Stick; (new) State: [10, 6, 6], [7]; Reward: 0''}. Figure~\ref{fig:prompt_depiction} presents a simplified depiction of the agent, while full examples are provided in Appendix~\ref{appendix: PromptRL-Agent}. State representations --  provided by the environment -- are often encoded in specific formats (e.g., data structures or numerical values such as hash values calculated out of structured representations). Therefore, decoding state representations can either be handled directly by the LLM -- requiring it to interpret the encoded state -- or delegated to an external function (e.g., a Python script)\footnote{Notably, the LLM can generate these scripts itself.}. Such script may include world knowledge on how to represent a particular state in a meaningful way in language.

\begin{figure}
  \includegraphics[width=\columnwidth]{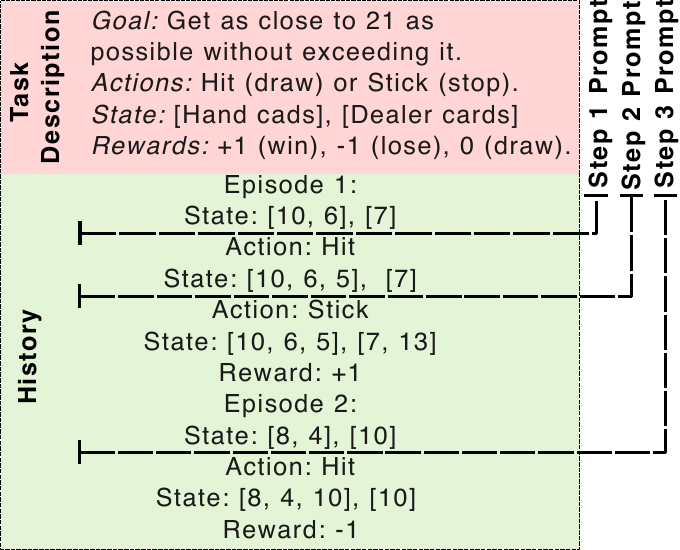}
  \caption
  {Example of a simplified prompt interaction for the Blackjack task,
    over three steps with the environment,
    equivalent to two episodes.
  }
  \label{fig:prompt_depiction}
\end{figure}

\subsection{PARL Training}
\label{subsection: Training}

Training a PARL agent requires creating the history $h$ between the agent and the environment to enable in-context learning for the LLM. Since LLMs do not retain the memory of previous outputs during inference, the prompt must explicitly accumulate information from prior steps and episodes.

Figure~\ref{fig:prompt_depiction} illustrates a simplified depiction of the prompt used for the blackjack RL task. In the first iteration (Step 1 Prompt), the input prompt contains only
the task description $\mathcal{T}$ (goal, action space, observation space, set of rewards), and the initial state. The LLM is prompted to generate an action from the specified action space (as described in Section~\ref{subsection: PromptRL-Agent}). This action (e.g., \textit{``Hit''}) is then fed to the environment, to produce a new state and rewards (e.g., \textit{``[10,6,5] and [7]''}, representing the three user’s and one dealer’s cards). These updates are concatenated with the $\mathcal{P}^{\text{PARL}}$ prompt to be used in the next step (e.g., Step 2 Prompt). This process iterates over $n$ steps, each time appending the log of interactions to the prompt (Eq.~5). With each new episode, additional logs are concatenated to the prompt, gradually increasing the context -- with more examples -- to enhance decision-making in future iterations through in-context learning.

\subsection{PARL Inference}
The inference process of a PARL agent uses the trained policy, represented as a prompt, which consists of two main components: (1) the task description (\(\mathcal{T}\), Section~\ref{subsection: PromptRL-Agent}), which includes the task goal, action space, state representation, and reward space, and (2) the history ($h = \{(\text{s}_1, \text{a}_1, \text{r}_1), \dots, (\text{s}_{n}, \text{a}_{n}, \text{r}_{n})\}$), which records the sequence of states, actions, and rewards with the environment. During inference, only the history from the current episode is concatenated at each step and removed once the episode ends.
The PARL agent uses the LLMs to generate an action from the defined action space.

\section{Experiments}
To address our research questions (Section~\ref{sec:intro}), we evaluate the PARL agent under three configurations (Section~\ref{sub:PromptRL-Agent Configurations}): \textit{full context}, \textit{context with random rewards}, and \textit{no context}. Each configuration is tested in two state-decoding variants (Section~\ref{sub:state_decoding}). GPT-4o\footnote{The total cost of the experiments was 145 USD. They have been performed in October 2024.} is used as the underlying language model for all experiments.

The experiments are conducted on three RL tasks: Blackjack, Frozen Lake, and Taxi (Section~\ref{sub: RL tasks}). These environments are non-linguistic, offering a clear contrast to typical LLM applications, where natural language is central.

\subsection{PARL Agent Configurations}
\label{sub:PromptRL-Agent Configurations}
\paragraph{Full History.}
This setup defines our standard approach for training PARL agents. We train a PARL agent for $100$ episodes for each task.
The history $h$ is gradually concatenated, meaning that episode $n$
includes context from all previous $n-1$ episodes.
For example, Episode $1$ has no context, while Episode $100$ has the history of all previous $99$ episodes ($h = \{(\text{s}_1, \text{a}_1, \text{r}_1), \dots, (\text{s}_{99}, \text{a}_{99}, \text{r}_{99})\})$.

\paragraph{History with Random Rewards.}
To assess whether the LLM learns from reward signals, we adopt a similar setup to the \textit{Full History}. The prompt includes the task description ($\mathcal{T}$) and interaction history ($h$), but with true rewards (\(\text{r}_t\)) replaced by random ones ($\text{r}_t^\text{rand}$), sampled uniformly from the reward set (\(\mathcal{R}\)). If performance remains comparable to setups with true rewards, it suggests the LLM is not learning from rewards but instead leveraging pre-trained knowledge.

\paragraph{No History.}
In this configuration, the PARL agent has no access to prior episode history. The prompt includes only the task description ($\mathcal{T}$) and the interaction history ($h$) from the current episode.
Specifically, \(h = \{(\text{s}_t, \text{a}_t, \text{r}_t)\}\)
contains only the states, actions, and rewards from the current episode ($t$).
Rewards ($\text{r}_t$) are included in the prompt only if provided by the environment at that
step (e.g., per-step vs. end-of-episode rewards).

\subsection{Reinforcement Learning Tasks}
\label{sub: RL tasks}

\paragraph{Blackjack.} The \textit{task} is to play a card game aiming to get closer to 21 than the dealer without exceeding it.
The \textit{state} is represented as a tuple: the player’s current sum, the value of the dealer’s visible card (1--10, where 1 represents an Ace), and whether the player holds a usable Ace (0 or 1). \textit{Actions} are \textit{``Hit''} (draw a card) or \textit{``Stick''} (end turn).
\textit{Rewards} are $+1$ for winning, $0$ for a tie, and $-1$ for losing.
The episode \textit{ends} if the player hits and the sum of the hand exceeds 21, or if the player sticks.
The \textit{objective} is to learn an optimal strategy for deciding when to \textit{``Hit''} or \textit{``Stick''}.\footnote{Gymnasium Blackjack description:
  \url{https://gymnasium.farama.org/environments/toy_text/blackjack/}.}

\paragraph{Frozen Lake.} The \textit{task} consists of crossing a frozen lake from start to goal without falling into any holes by walking over the frozen lake.
The \textit{state} represents the player’s position as a single integer, given by $\textrm{row}_{t} \cdot \textrm{ncols} + \textrm{col}_{t}$, where $\textrm{row}_{t}$ and $\textrm{col}_{t}$ are the current row and column, and $\textrm{ncols}$ is the total number of columns.
\textit{Actions} are of moving \textit{Up}, \textit{Down}, \textit{Left}, or \textit{Right}. \textit{Rewards} are $+1$ for reaching the goal and $0$ otherwise.
The episode \textit{ends} if the player moves into a hole, reaches the goal, or exceeds a maximum
number of movements.
The \textit{objective} is to learn an optimal path to the goal while minimizing the risk of falling
into holes.\footnote{Gymnasium Frozen Lake description:
  \url{https://gymnasium.farama.org/environments/toy_text/frozen_lake/}.}

\paragraph{Taxi.} The \textit{task} involves driving a passenger in a grid world, picking them up, and dropping them at one of the four locations on the map, one at a time.
The \textit{state} is encoded as a single integer that tracks the taxi’s position, the passenger’s location, and the destination: $((\textrm{taxi}_{\textrm{row}} \cdot 5 + \textrm{taxi}_{\textrm{col}}) \cdot 5 + \textrm{passenger}_{\textrm{location}}) \cdot 4 + \textrm{passenger}_{\textrm{destination}}$.
\textit{Actions} consist of \textit{Down}, \textit{Up}, \textit{Left}, \textit{Right}, \textit{Pickup} or \textit{Drop-off}. \textit{Rewards} are $+20$ for \textit{successfully dropping off} a passenger at the correct destination, $-1$ for each \textit{step taken}, and $-10$ for attempting illegal \textit{Pickup} or \textit{Drop-off} actions.
The episode \textit{ends} when the taxi successfully drops off the passenger at the correct
destination or when the maximum number of steps is reached.
The \textit{objective} is to learn efficient pick-up and drop-off strategies to minimize steps and
maximize rewards.\footnote{Gymnasium Taxi task description: \url{https://gymnasium.farama.org/environments/toy_text/taxi/}.}
\subsection{State-decoding Variants}
\label{sub:state_decoding}

Two configurations are used for the state ($s$) in the history $h$: (1) the environment state is passed directly to the prompt without preprocessing, requiring the LLM to interpret the raw symbolic or numeric form. (2) An external decoding function (a Python script unique to each environment) converts the state into a natural-language description. In \textbf{Blackjack}, the raw state is used as produced by the environment, e.g., \textit{``State: [10, 6, 4], [7]''}, while the decoded version expresses it in natural language, e.g., \textit{``The player's hand totals 20, and the dealer shows a 7.''}. In \textbf{Frozen Lake}, the raw state is a grid index, e.g., \textit{``State: 6''}, and the decoded version describes the position, e.g., \textit{``The player is currently located at row 1, column 2 in a 4x4 grid.''}. In \textbf{Taxi}, the raw state is a tuple encoding position and passenger data, e.g., \textit{``State: 6''}, and the decoded version reformulates it as \textit{``The taxi is at row 0, column 0. The passenger is at location Green, and the destination is Yellow.''}.

\begin{figure*}[h]
  \centering
  \begin{subfigure}[b]{0.32\textwidth}
    \centering
    \includegraphics[width=\textwidth]
    {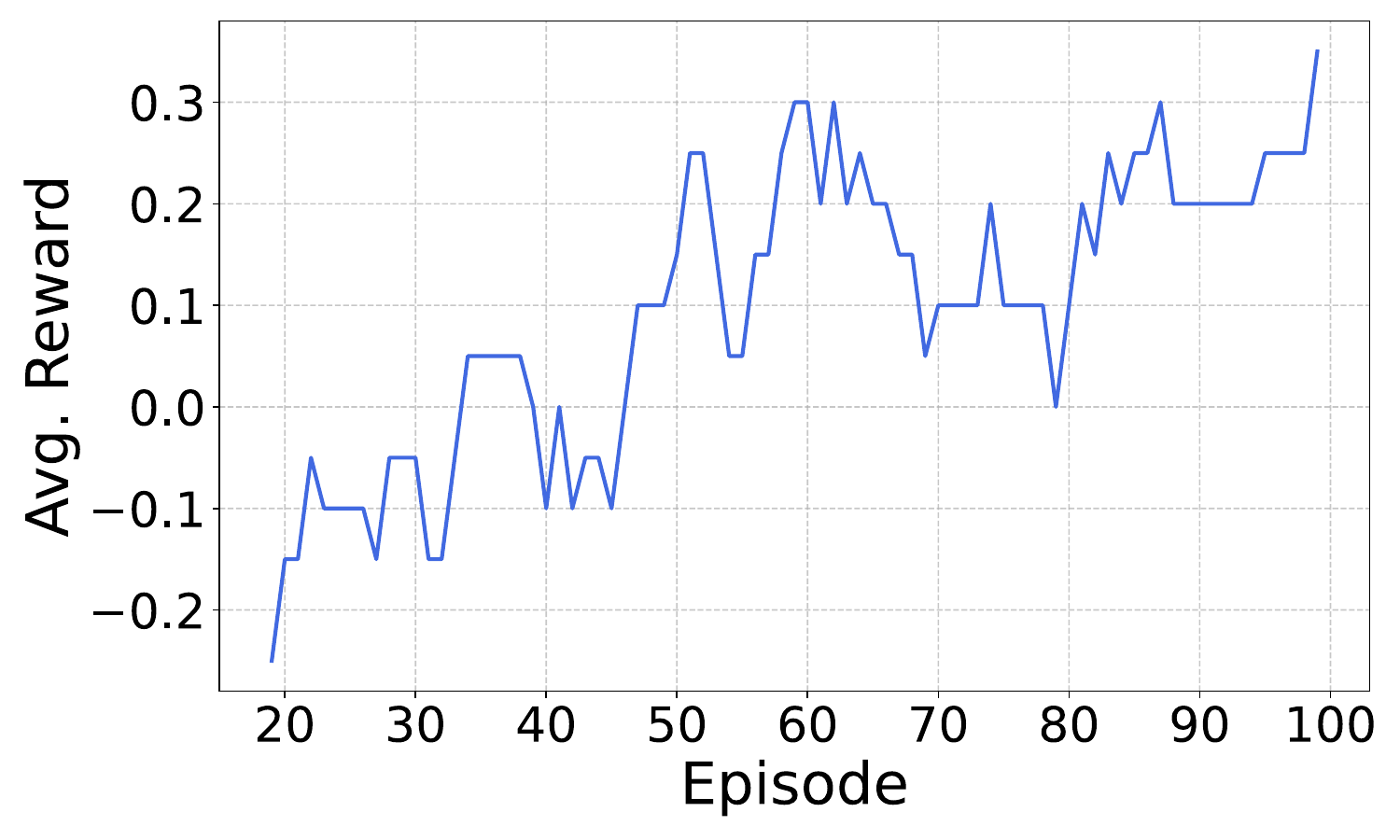}
    \caption{Blackjack}
    \label{fig:lC_blackjack}
  \end{subfigure}
  \hfill
  \begin{subfigure}[b]{0.32\textwidth}
    \centering
    \includegraphics[width=\textwidth]
    {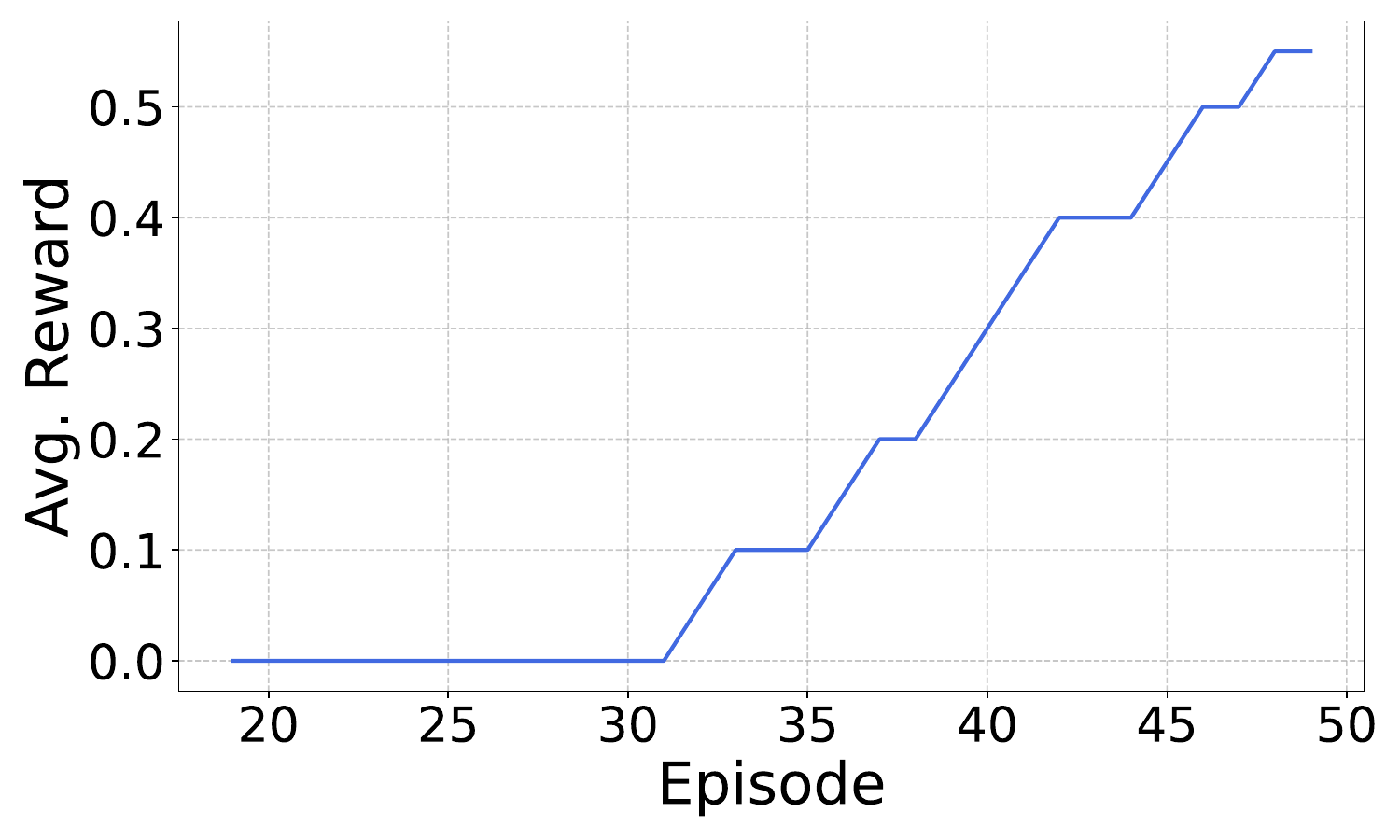}
    \caption{Frozen Lake}
  \end{subfigure}
  \hfill
  \begin{subfigure}[b]{0.32\textwidth}
    \centering
    \includegraphics[width=\textwidth]{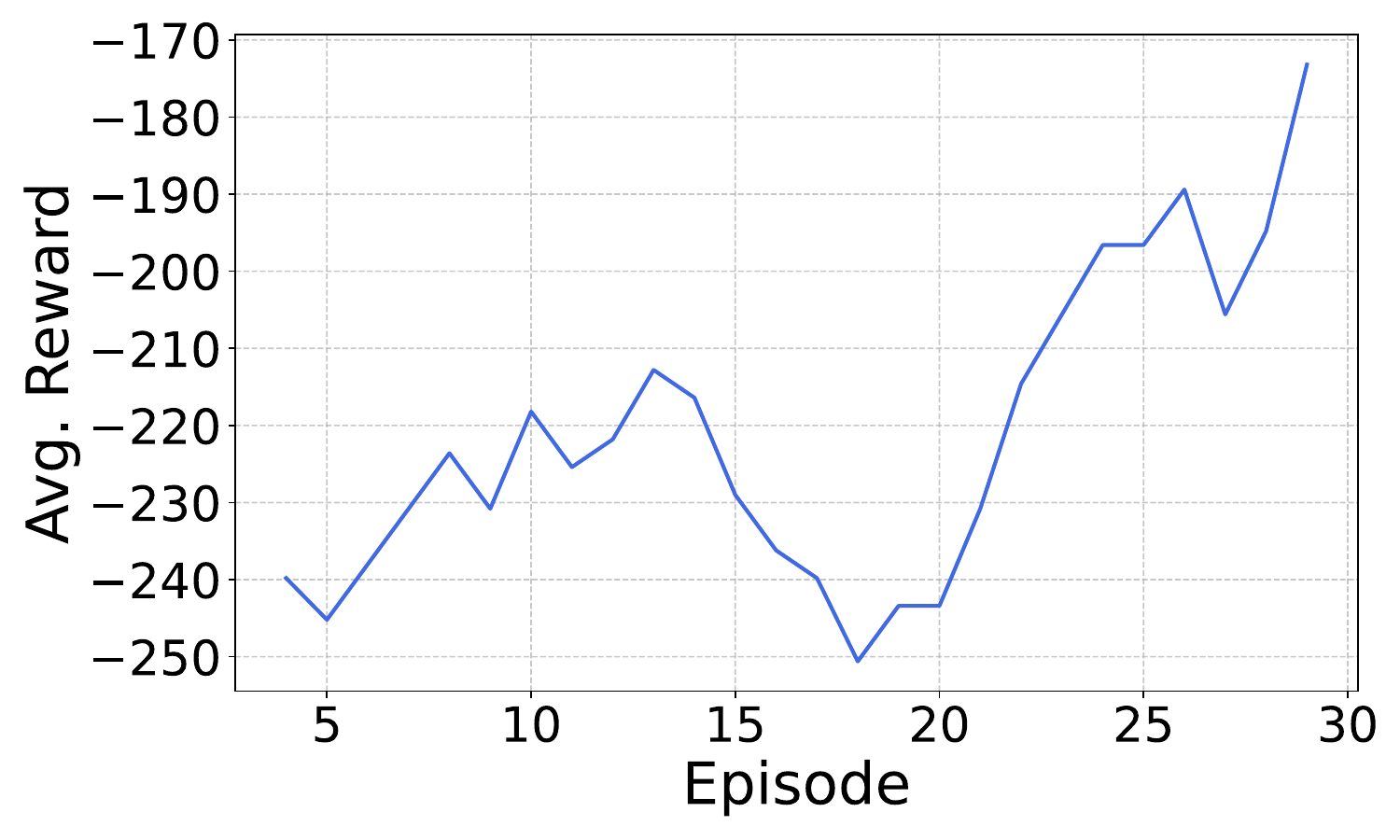}
    \caption{Taxi}
  \end{subfigure}
  \caption{
    Average reward during training with the PARL agent using LLM self-decoding of states. A smoothing window of five episodes is applied.}
  \label{fig:learning_curves_llm_self}
\end{figure*}

\begin{figure*}[h]
  \centering
  \begin{subfigure}[b]{0.32\textwidth}
    \centering
    \includegraphics[width=\textwidth]
    {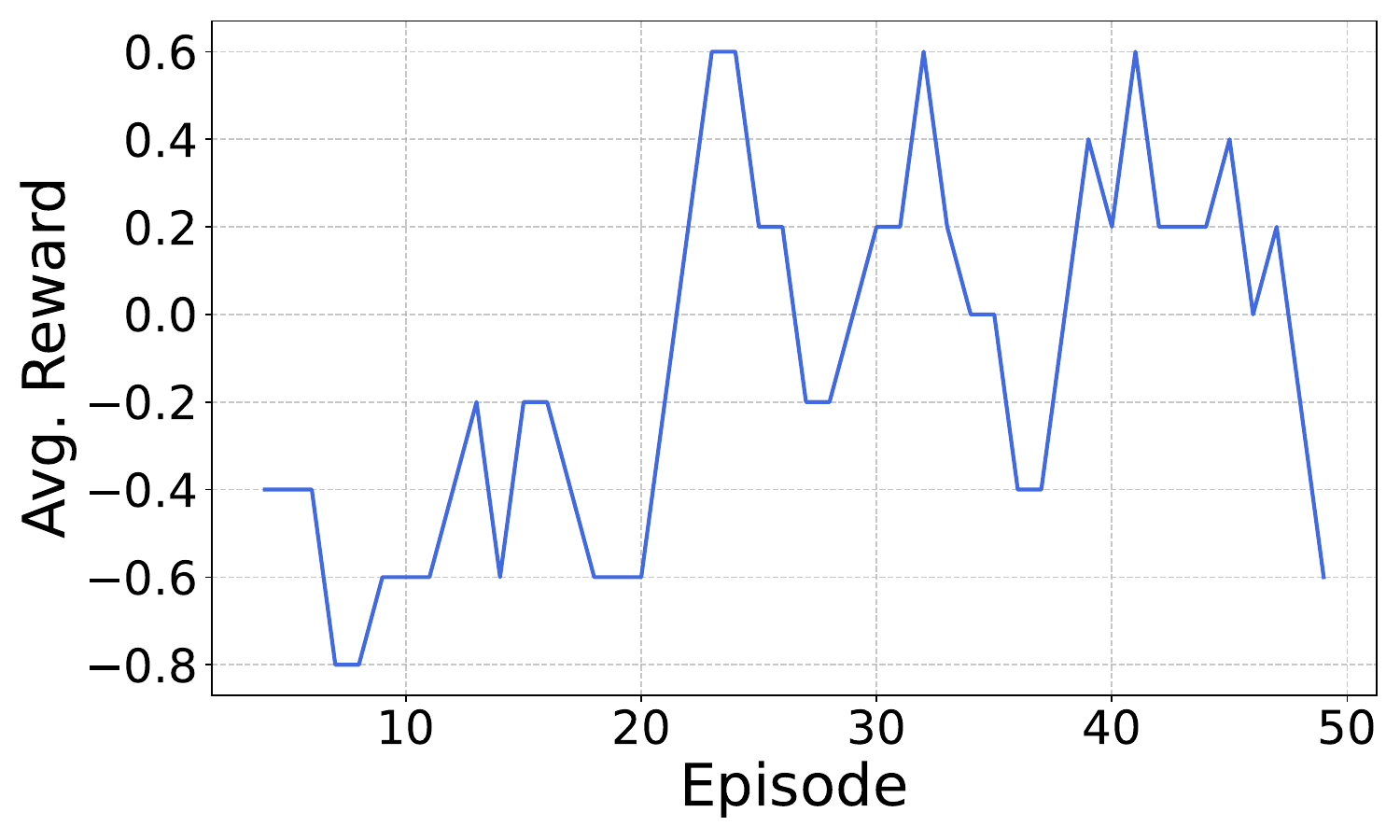}
    \caption{Blackjack}
  \end{subfigure}
  \hfill
  \begin{subfigure}[b]{0.32\textwidth}
    \centering
    \includegraphics[width=\textwidth]
    {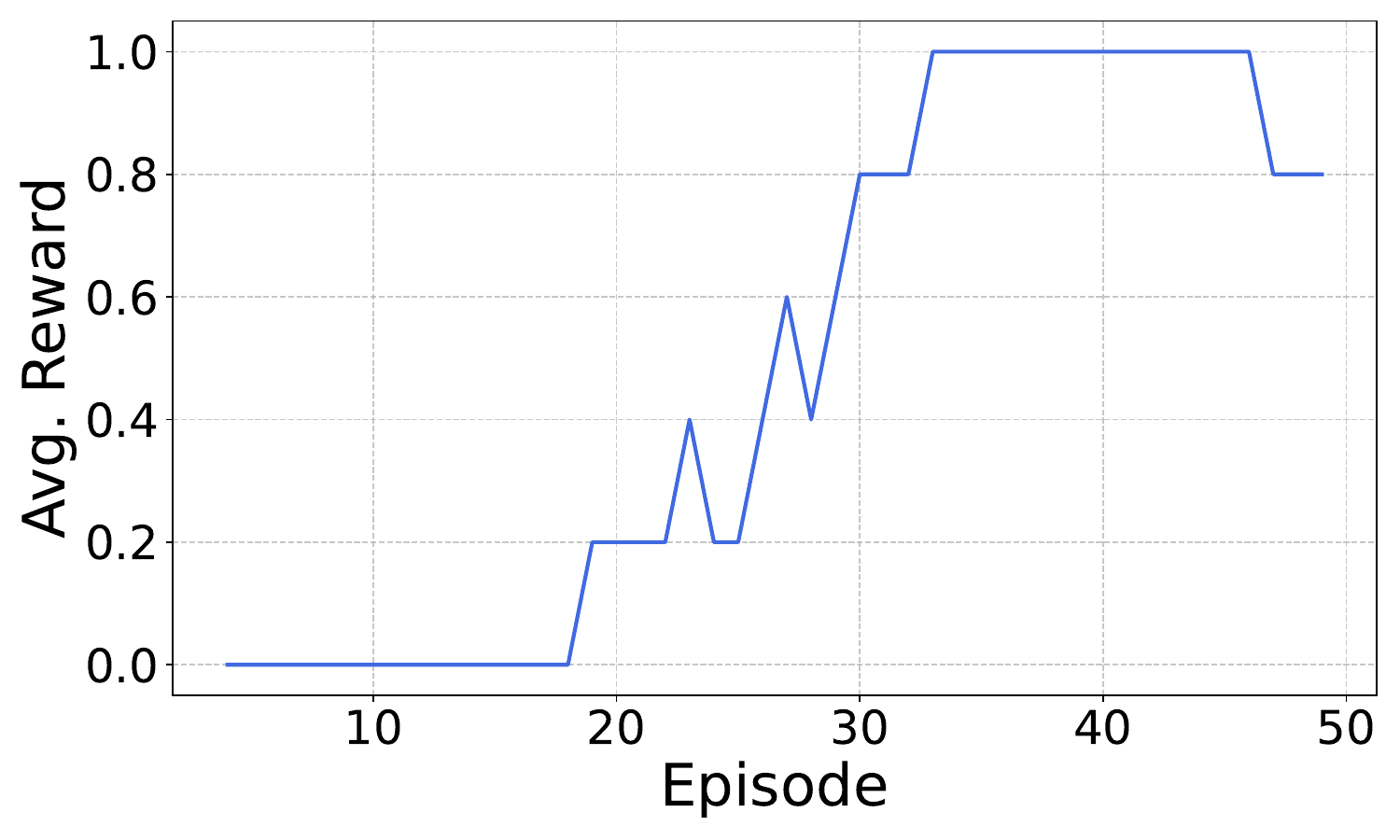}
    \caption{Frozen Lake}
  \end{subfigure}
  \hfill
  \begin{subfigure}[b]{0.32\textwidth}
    \centering
    \includegraphics[width=\textwidth]{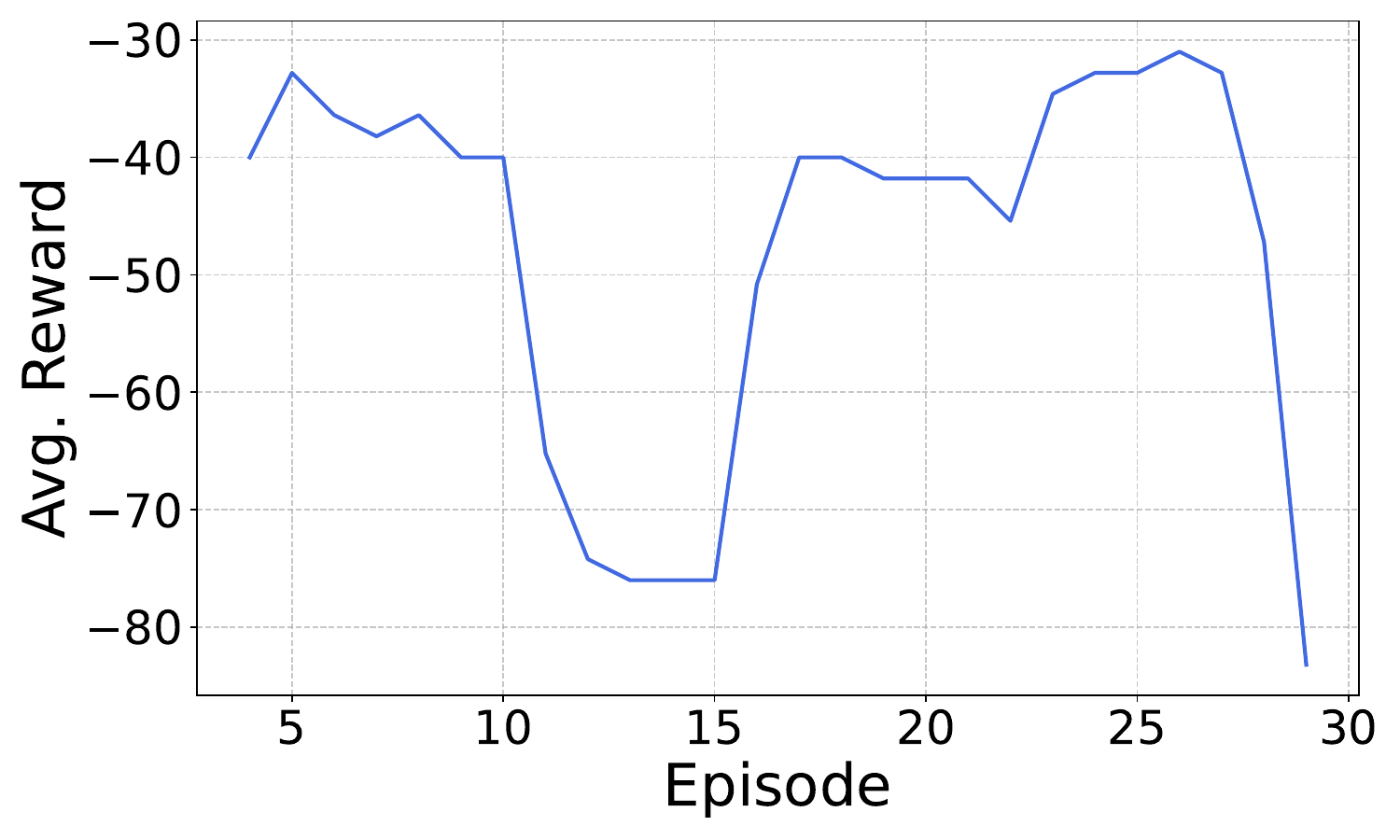}
    \caption{Taxi}
  \end{subfigure}
  \caption{Average reward during training with the PARL agent using script-based state decoding. A smoothing window of five episodes is applied.}
  \label{fig:learning_curves_llm}
\end{figure*}

\begin{figure*}[h!]
  \centering
  \begin{subfigure}[b]{0.32\textwidth}
    \centering
    \includegraphics[width=\textwidth]
    {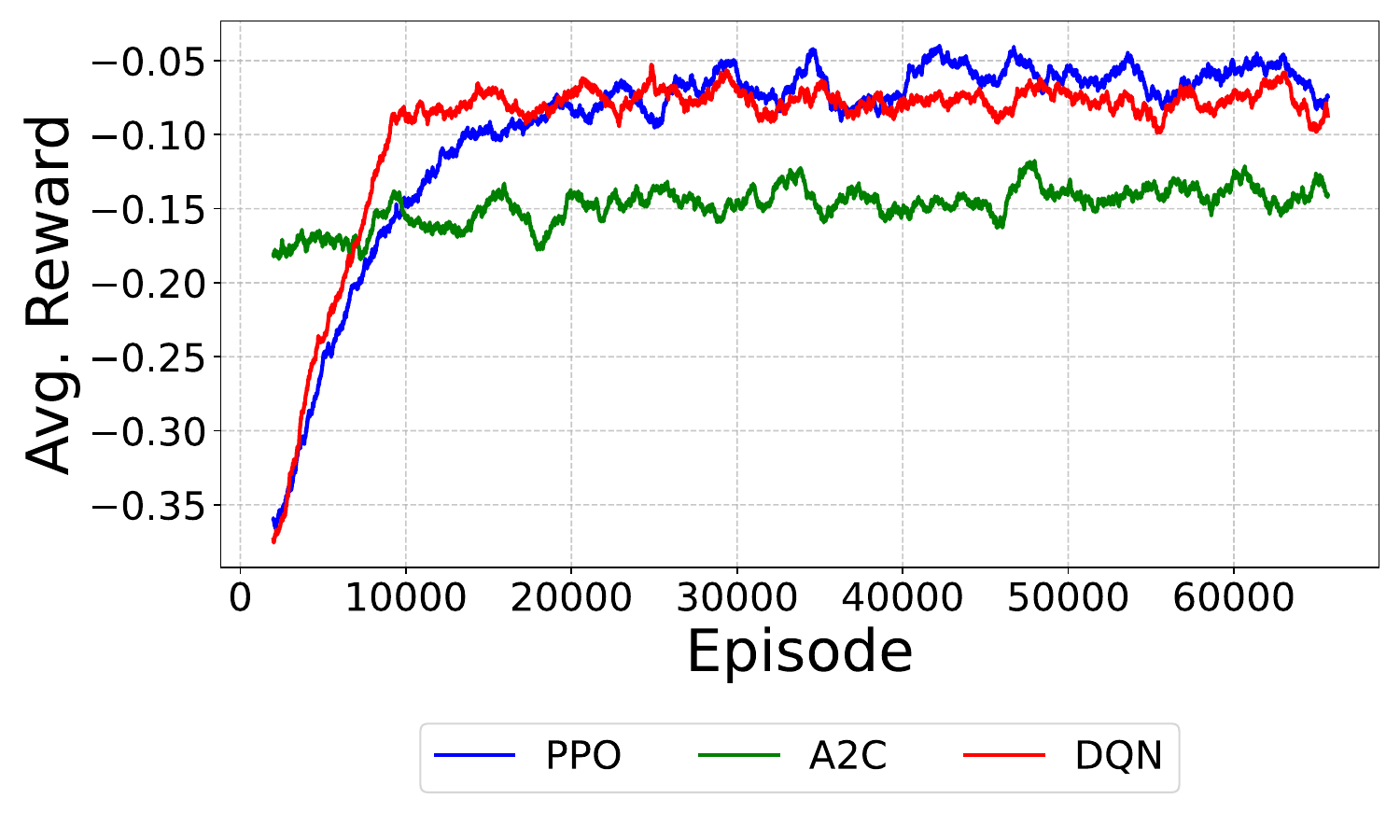}
    \caption{Blackjack}
    \label{fig:plot1}
  \end{subfigure}
  \hfill
  \begin{subfigure}[b]{0.32\textwidth}
    \centering
    \includegraphics[width=\textwidth]
    {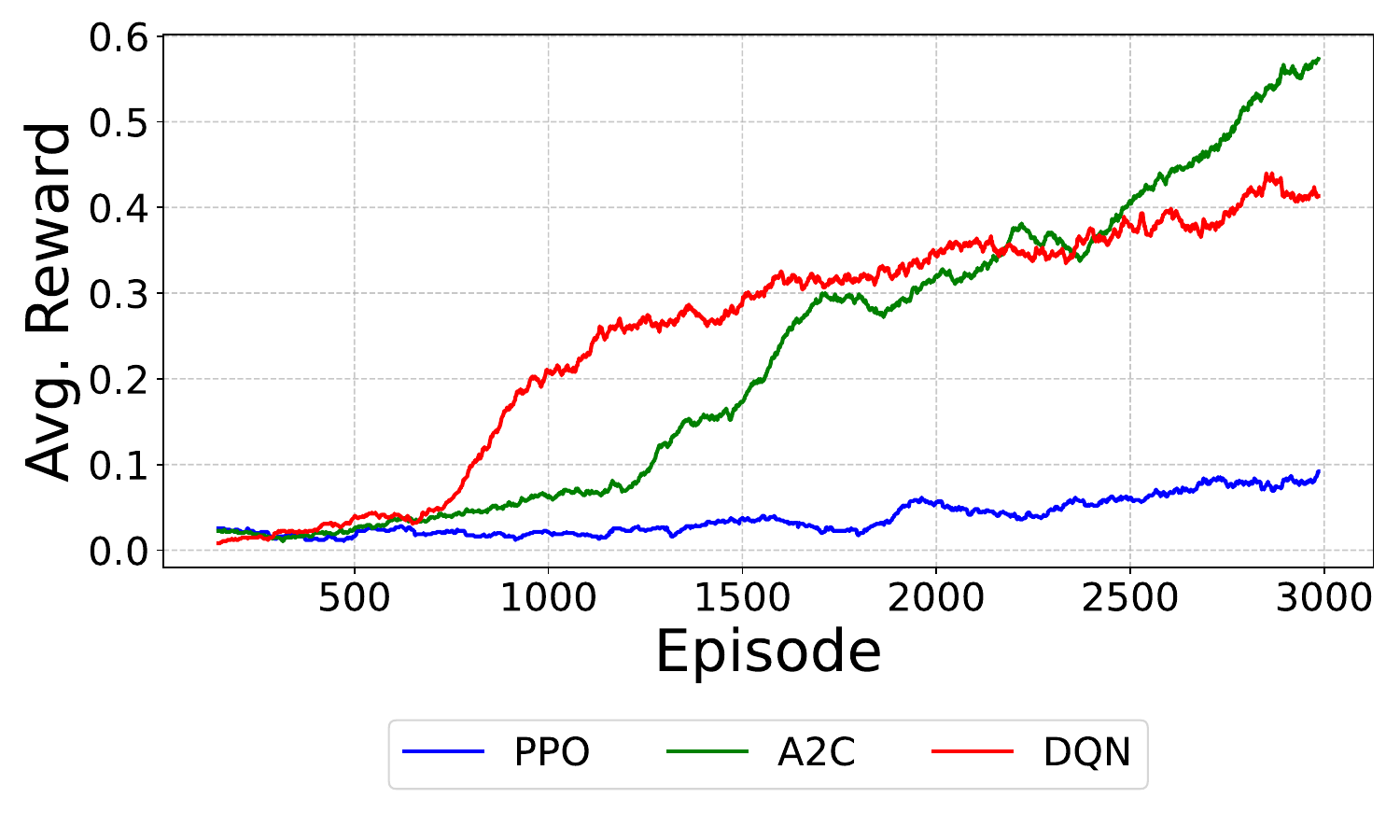}
    \caption{Frozen Lake}
  \end{subfigure}
  \hfill
  \begin{subfigure}[b]{0.32\textwidth}
    \centering
    \includegraphics[width=\textwidth]{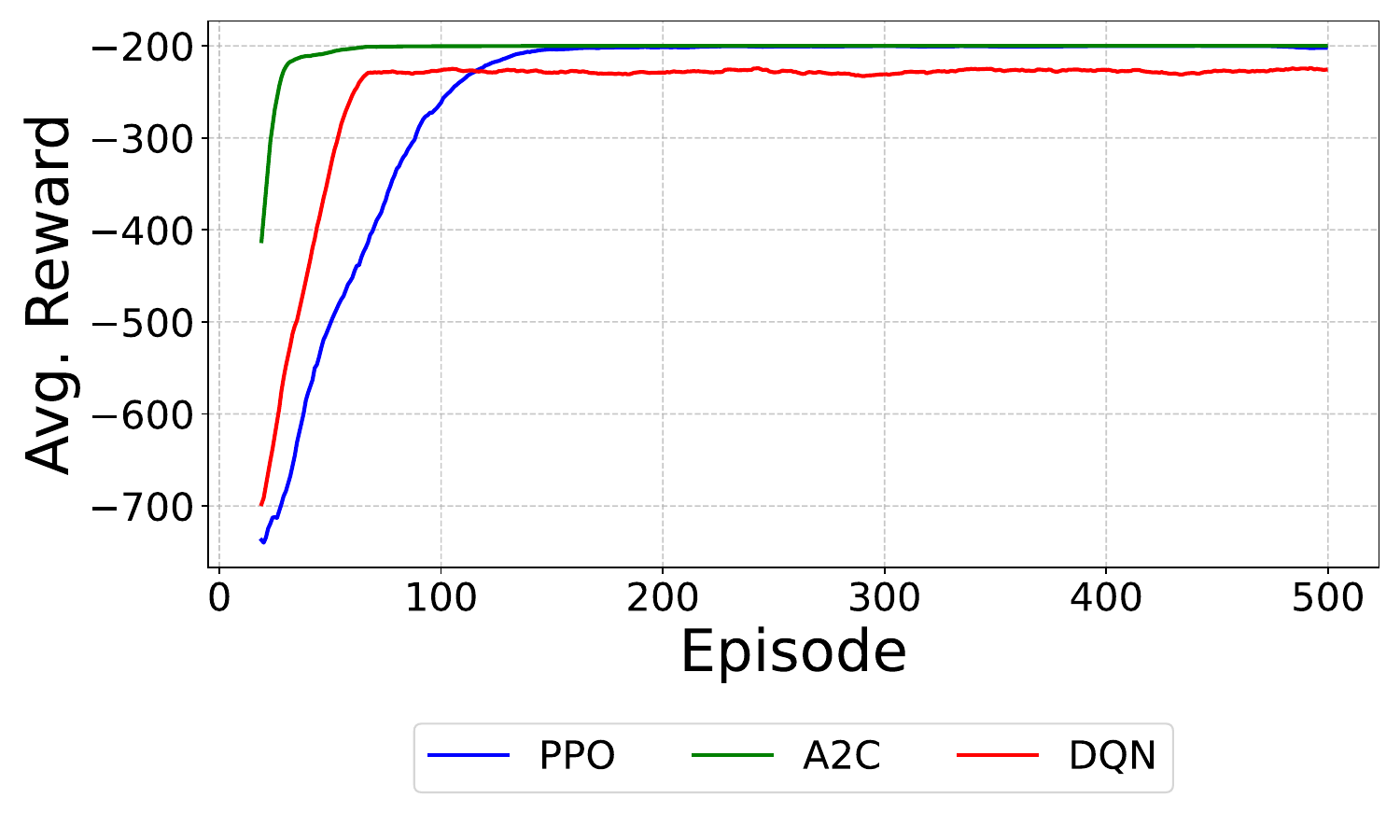}
    \caption{Taxi}
  \end{subfigure}
  \caption{Comparison of average training rewards for SOTA agents (smoothing window = 200).
  }
  \label{fig:learning_curves_rewards_sota}
\end{figure*}

\begin{figure*}[h!]
  \centering
  \begin{subfigure}[b]{0.32\textwidth}
    \centering
    \includegraphics[width=\textwidth]{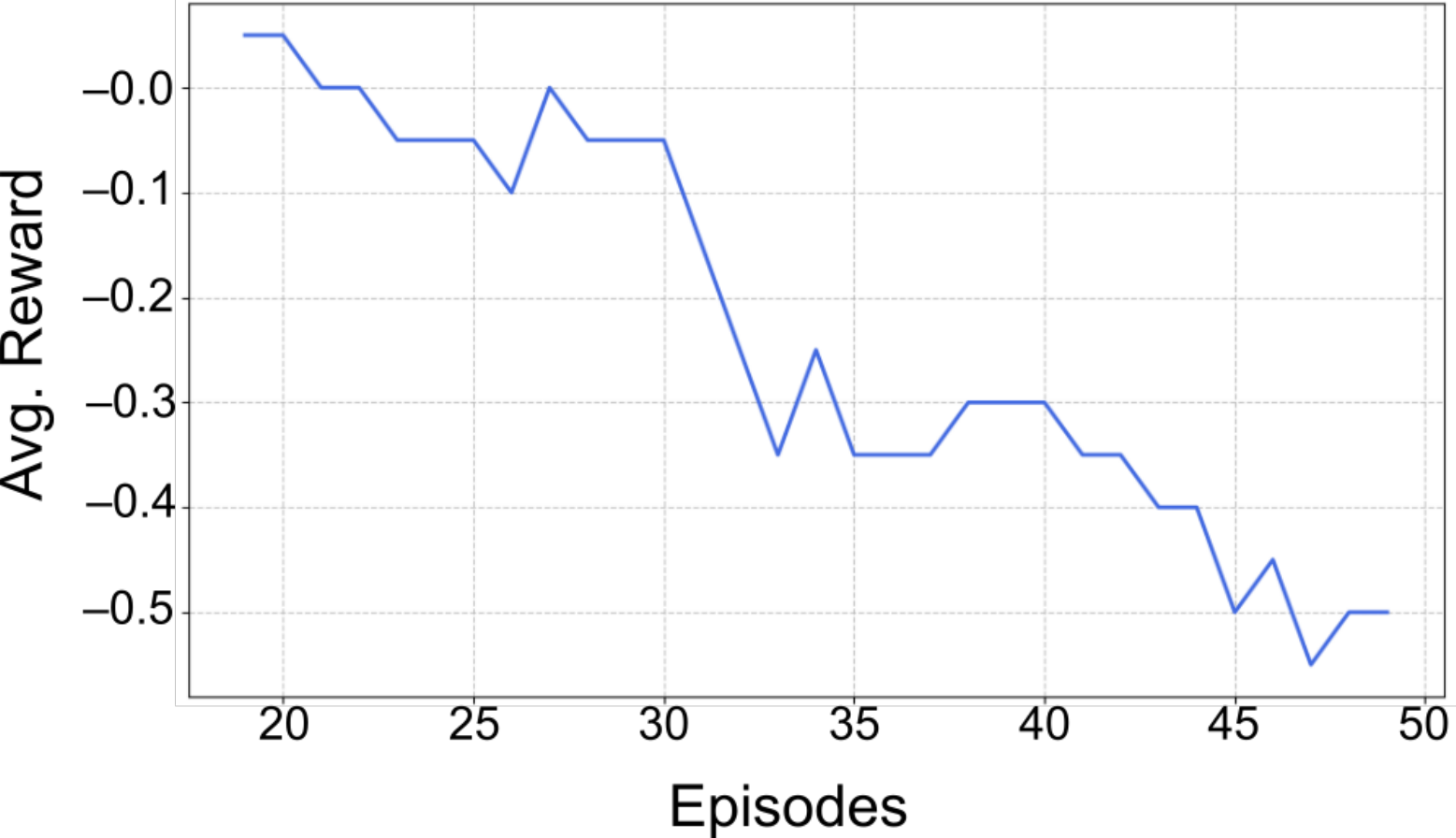}
    \caption{Blackjack}
  \end{subfigure}
  \hfill
  \begin{subfigure}[b]{0.32\textwidth}
    \centering
    \includegraphics[width=\textwidth]
    {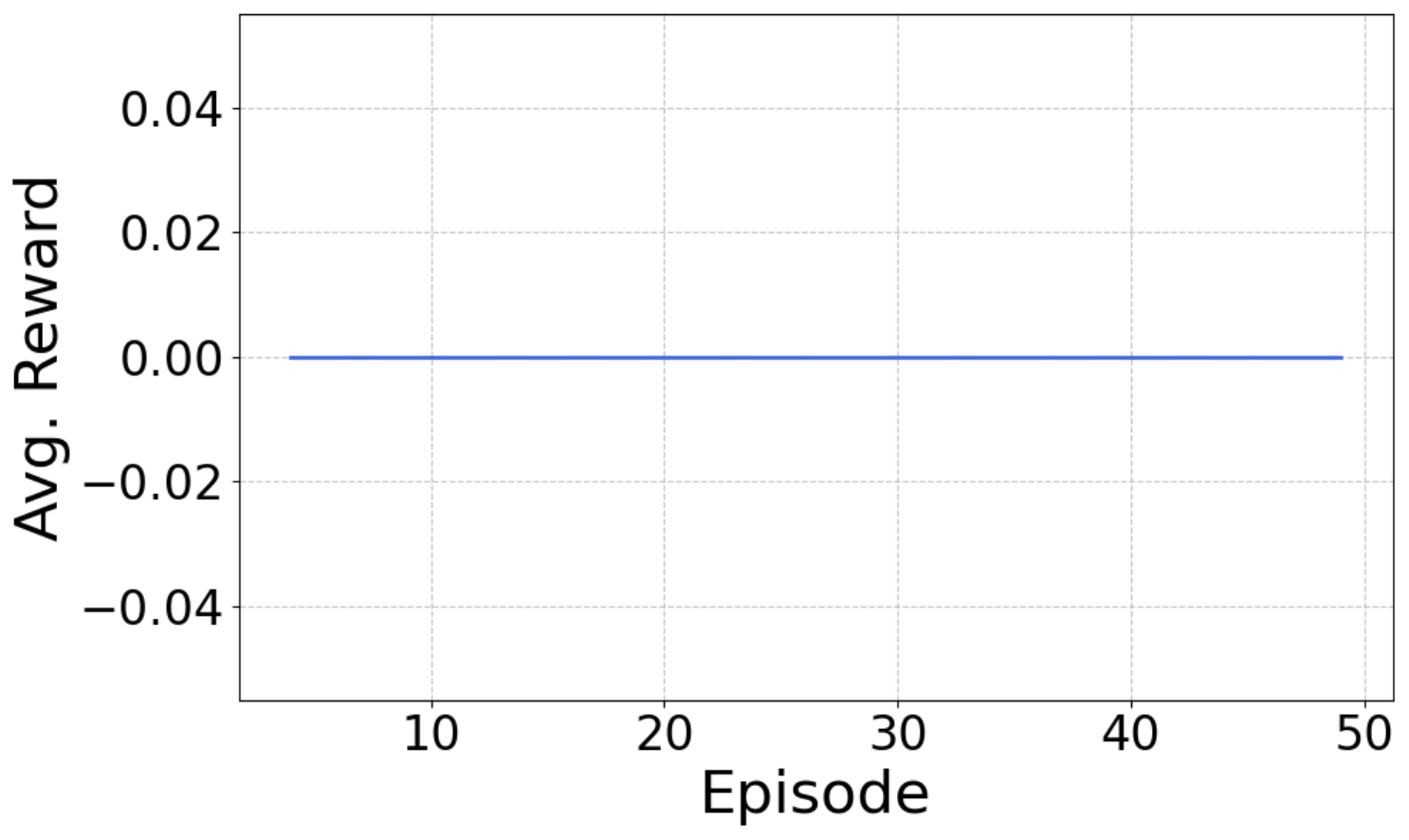}
    \caption{Frozen Lake}
  \end{subfigure}
  \hfill
  \begin{subfigure}[b]{0.32\textwidth}
    \centering
    \includegraphics[width=\textwidth]{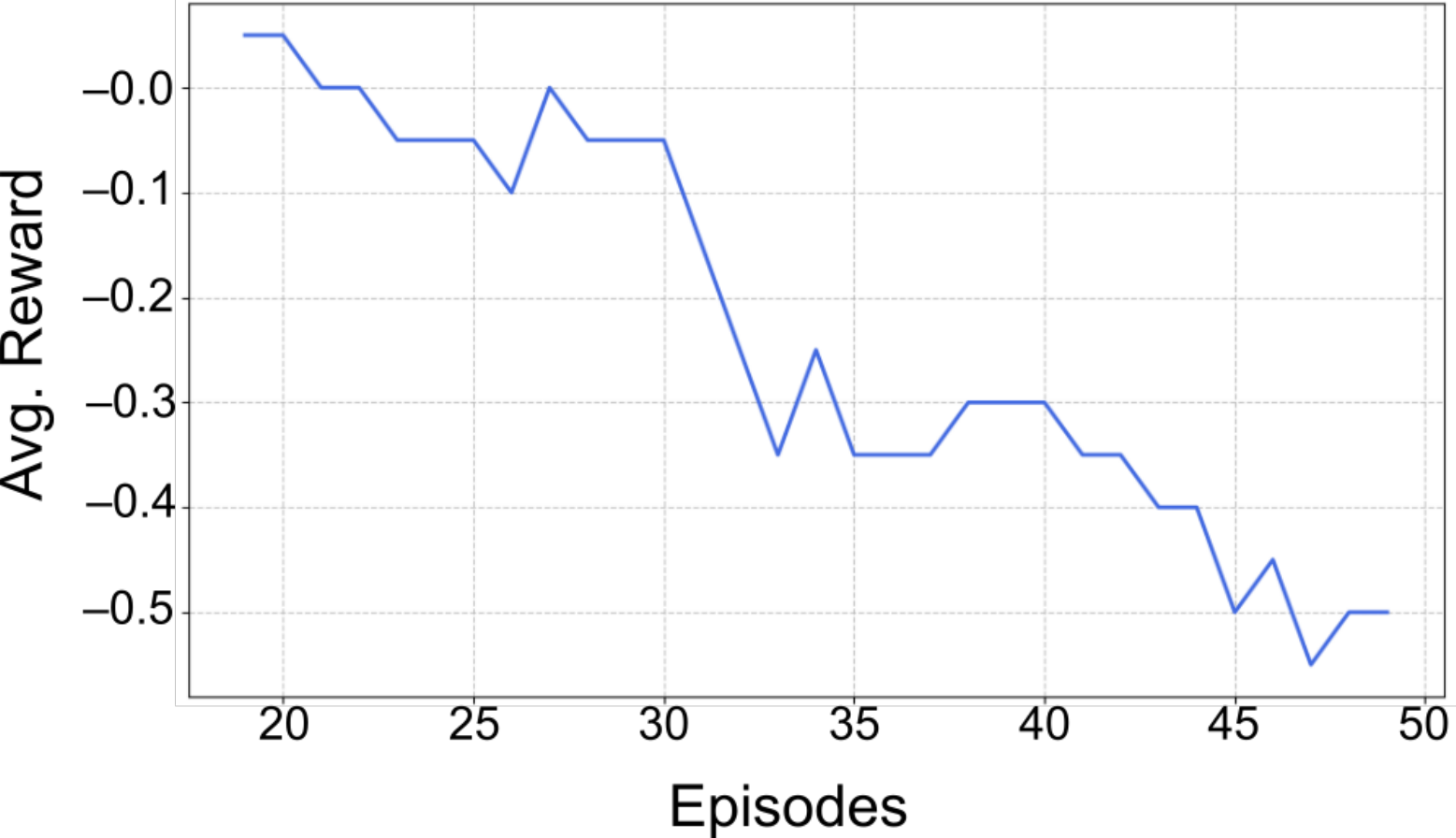}
    \caption{Taxi}
    \label{fig: learning curves fake rewards}
  \end{subfigure}
  \caption
  {Comparison of the episode real reward during PARL agent training when informing the agent with random or incorrect rewards.}
  \label{fig:learning_curves_llm_fake}
\end{figure*}

\section{Results}
\label{sec:results}

Before discussing the three configurations of the PARL agent used to address our research questions, we first examine the two state-decoding variants by comparing their training curves: LLM self-decoding (Figure~\ref{fig:learning_curves_llm_self}) and script-based decoding (Figure~\ref{fig:learning_curves_llm}). Note that reward values are only comparable between the two decoding variants within the same environment, as each task defines its own reward scale and penalty structure (see Section~\ref{sub: RL tasks} and Appendix~\ref{appendix:rl_tasks} for detailed task descriptions).

In Blackjack, the script-based decoding reaches an average reward of 0.6 after only 23 episodes, while the self-decoding only reaches 0.3 after 50 episodes. A similar pattern can be seen in Frozen Lake, where the script variant achieves 1 after 32 episodes, whereas the self-decoding only reaches 0.55 at the end of training. The Taxi environment is more challenging due to its more complex encoding. Here, the LLM is unable to reliably decode positions on its own, leading to random actions. In contrast, the script provides accurate state representations, allowing the LLM to learn low-penalty actions.

Overall, the script-based decoding shows faster learning and higher performance, with steeper learning curves. Based on these results, the following section focuses only on the script-based decoding method.

\subsection{RQ1: Can a PARL Agent Learn from Environment Interactions?}
\label{sub:RQ1}

To evaluate whether the PARL agent learns to take better actions
based on previous episodes, we investigate three subquestions:
\textit{RQ1.1: Does the PromptRL-Agent benefit from contextual examples during training and inference?}
\textit{RQ1.2: What happens when random or incorrect rewards are provided?}; and \textit {RQ1.3: How does the PARL policy compare to state-of-the-art policies?}

\paragraph
{RQ 1.1: Does a PARL agent benefit from contextual examples during training and inference?}
We start by analyzing the role of context during training and inference (Exp.~1; PARL with full history).
Figure~\ref{fig:learning_curves_llm} shows the learning curves for the PARL agent across the three RL tasks.
The agent benefits from the contextual examples provided, suggesting that the LLM can learn to make better action choices using contextual examples.
In Blackjack, the model’s average reward increases from $-0.8$ to $0.6$ \cite[human performance is around $0.48$; ][]{Carlin_Robinson_2009}.

A similar behavior is observed in the Frozen Lake task, where the model initially explores to find the correct path.
This training improvement shows the agent's ability to learn better action policies based on the
history $h$.
The Taxi task presents a greater challenge with its larger action space.
The LLM is unable to learn to solve the task.

Table~\ref{table:Average rewards} reports the average rewards obtained by the three PARL agent configurations during inference.
It presents the average rewards over $100$ evaluation episodes for the three configurations of the PARL agent (top) and three state-of-the-art agents (bottom).
When comparing the \textit{Full History} configuration to the \textit{History with Random Rewards} and \textit{No History} setups, we observe that history improves the PARL agent's performance.
For instance, in Blackjack, the PARL agent achieves an average reward of $0.2$ with \textit{Full History}, outperforming the \textit{History with Random Rewards} ($-0.12$) and \textit{No History} ($-0.09$) setups.
Similarly, in the Frozen Lake task, the \textit{Full History} setup achieves an average reward of $0.66$, compared to the $0.0$ from the other configurations.
On the Taxi task, the PARL agent with \textit{Full History} presents a similar behavior to the training. The agent fails to solve the task but tries to minimize negative rewards ($-696.6$ compared to $-807.5$ for \textit{History with Random Rewards} and $-568.95$ for \textit{No History}).
The agent frequently selects sub-optimal actions, such as illegal pick-up or drop-off, which incur a significant penalty ($-10$).

\begin{table*}[]
  \fontsize{8}{11}\selectfont
  \setlength{\tabcolsep}{5pt}
  \begin{tabular}{l|rr|rr|rr}
    \toprule
    \multirow{2}{*}{\textbf{Setup}} &
                                      \multicolumn{2}{c}{\textbf{Blackjack}} &
                                                                               \multicolumn{2}{c}{\textbf{Frozen Lake}} &
                                                                                                                          \multicolumn{2}{c}{\textbf{Taxi}}\\
                                    & \textbf{Avg. Reward} & \textbf{Avg. Length} & \textbf{Avg. Reward}
                                    & \textbf{Avg. Length} & \textbf{Avg. Reward} & \textbf{Avg. Length}
    \\ \toprule

    PARL Full History   & 0.20 (.94)           & 1.5 (.66)           &
                                                                       0.66 (.47)           & 6.37 (.89)           & $-$
                                                                                                                     696.6 (28.23)     & 100 (0.0)            \\
    PARL Random Rewards & $-$ 0.12 (.94)        & 1.5 (.69)           & 0 (0)                &
                                                                                               3.86 (.46)           & $-$ 807.5 (103.11)    & 100 (0.0)
    \\
    PARL No History     & $-$
                          0.07 (.93)        & 1.7 (.80)           & 0 (0)                &
                                                                                           3.34 (.97)           & $-$ 568.9 (64.95)    & 100 (0.0)
    \\
    \midrule
    PPO                & 0.04 (.96)           & 1.47 (.64)           &
                                                                       0.67 (.47)           & 34.69 (23.69)        & $-$
                                                                                                                     100 (0.0)         & 100 (0.0)            \\
    DQN                & $-$
                         0.04 (.95)        & 1.56 (.74)           & 0.61 (.48)           &
                                                                                           36.84 (26.06)        & $-$ 100 (0.0)         & 100 (0.0)
    \\
    A2C                & $-$
                         0.02 (.92)        & 1.24 (.58)           & 0.62 (.48)           &
                                                                                           34.96 (23.56)        & $-$ 100 (0.0)         & 100 (0.0)
    \\
    \bottomrule
  \end{tabular}
  \caption{Average (Avg.)
    reward and episode length across 100 episodes for the Blackjack, Frozen Lake, and Taxi tasks, comparing three PARL agent configurations (top) with state-of-the-art
    (SOTA; bottom) agents. Standard deviation is in parentheses.}
  \label{table:Average rewards}
\end{table*}

\paragraph{RQ1.2: What happens when random or incorrect rewards are provided?}

In the configuration where random rewards are included as context, the model fails to learn effectively across all tasks.
Figure~\ref{fig:learning_curves_llm_fake} illustrates that average rewards remain low regardless of training.
In Blackjack, the agent's average reward decreases to $-0.5$.
In Frozen Lake, the agent fails to learn a policy, achieving an average reward of $0.0$ (not succeeding in the task even once).
In the Taxi task, the agent selects random actions, accumulating high penalties, likely due to the large state and action space.
This confirms that irrelevant or misleading context introduces noise, preventing the model from identifying optimal actions.

\begin{figure*}[h]
  \centering
  \begin{subfigure}[b]{0.32\textwidth}
    \centering
    \includegraphics[width=\textwidth]
    {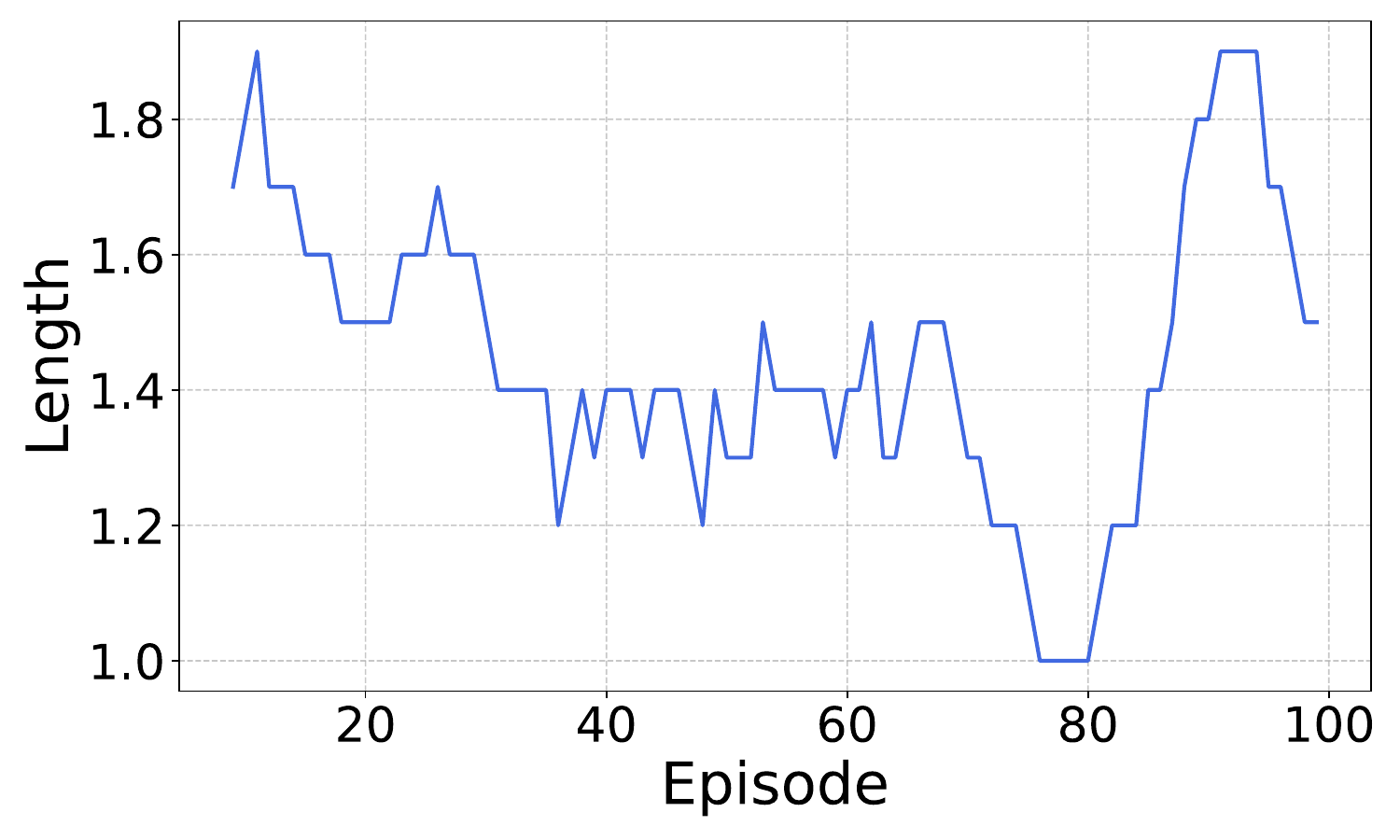}
    \caption{Blackjack}
    \label{fig:Learning curve llm}
  \end{subfigure}
  \hfill
  \begin{subfigure}[b]{0.32\textwidth}
    \centering
    \includegraphics[width=\textwidth]
    {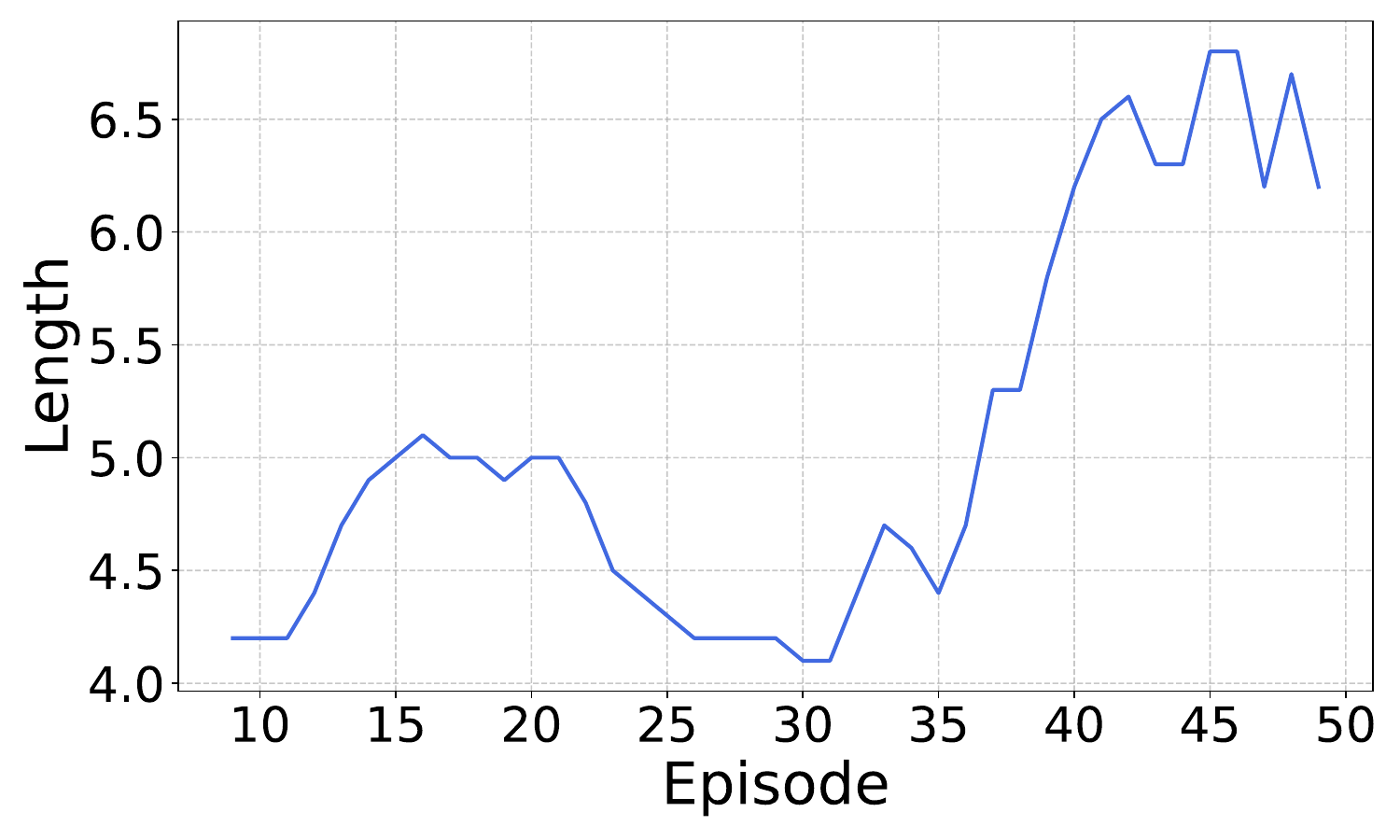}
    \caption{Frozen Lake}
    \label{fig:episode length frozen lake} 
  \end{subfigure}
  \hfill
  \begin{subfigure}[b]{0.32\textwidth}
    \centering
    \includegraphics[width=\textwidth]{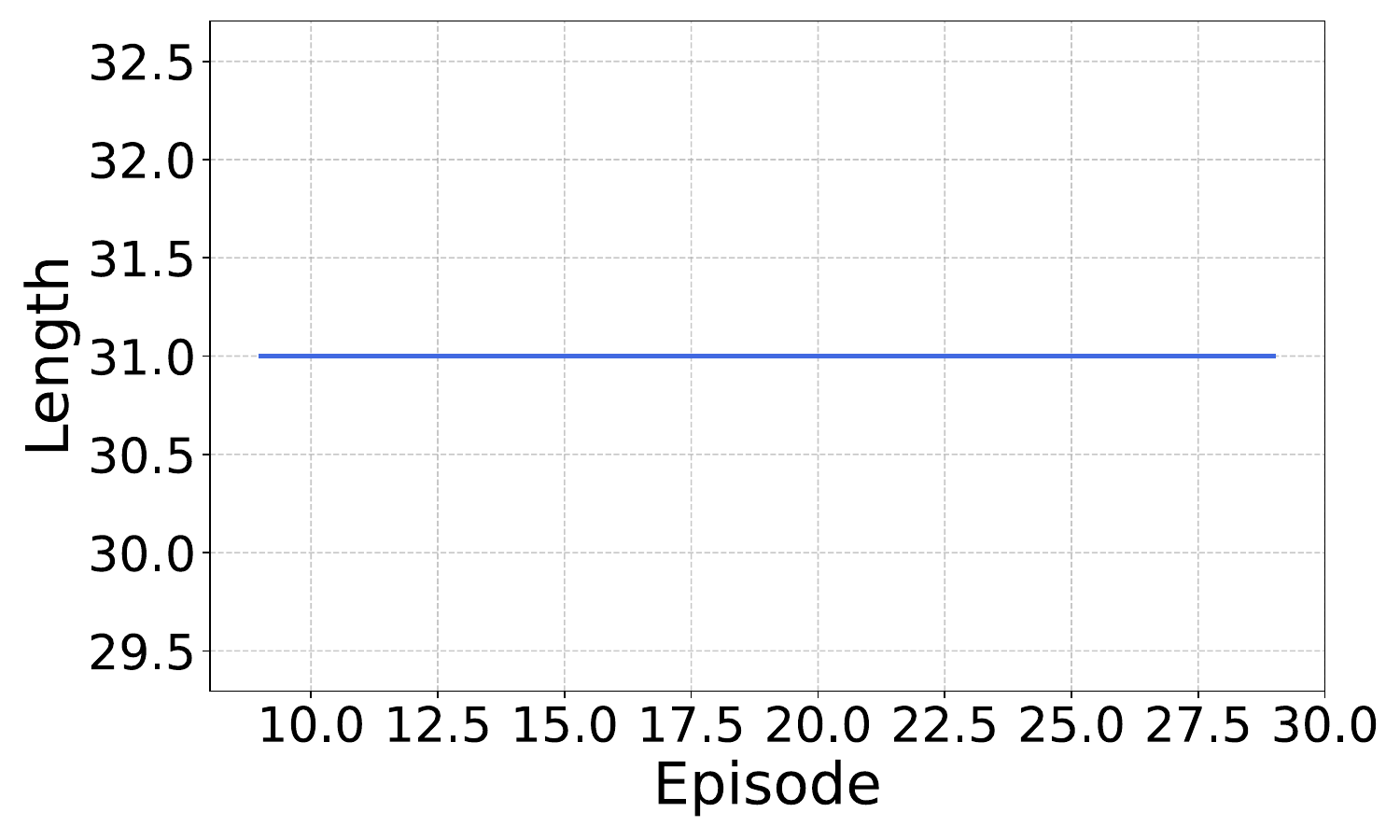}
    \caption{Taxi}
  \end{subfigure}
  \caption{Average episode length during training from PARL agent,
    with a smoothing window size of five.}
  \label{fig:episode length}
\end{figure*}

\begin{figure*}[h!]
  \centering
  \begin{subfigure}[b]{0.32\textwidth}
    \centering
    \includegraphics[width=\textwidth]
    {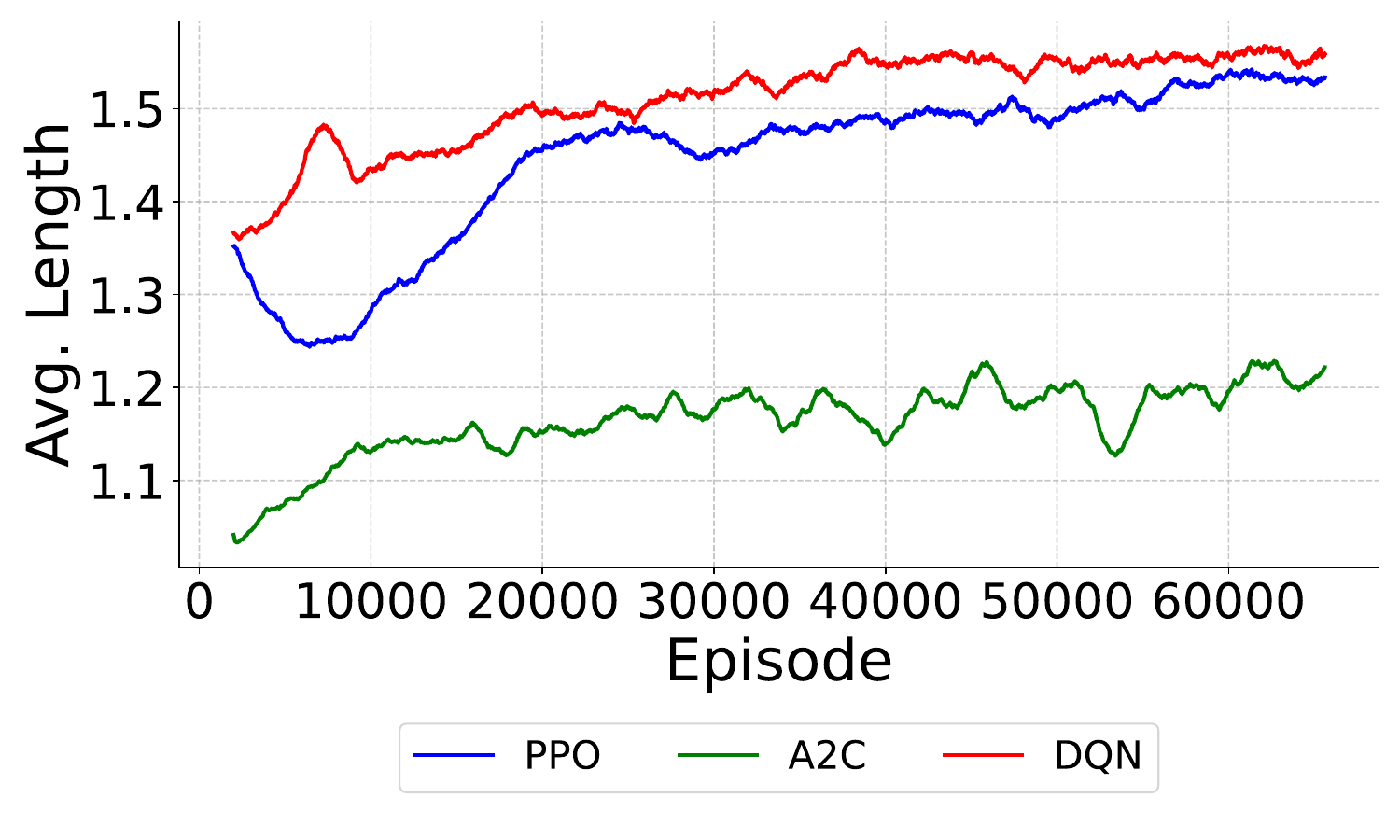}
    \caption{Blackjack}
  \end{subfigure}
  \hfill
  \begin{subfigure}[b]{0.32\textwidth}
    \centering
    \includegraphics[width=\textwidth]
    {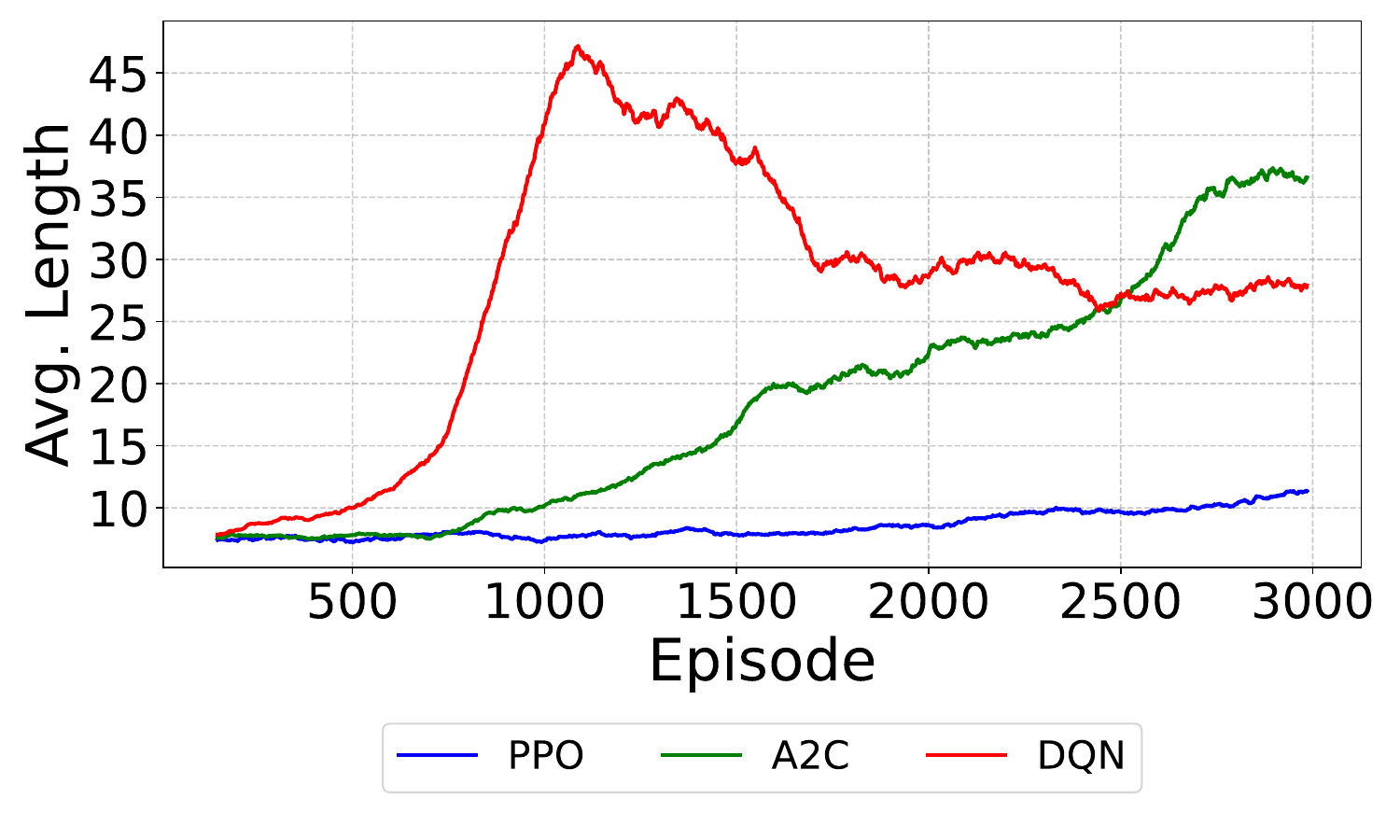}
    \caption{Frozen Lake}
  \end{subfigure}
  \hfill
  \begin{subfigure}[b]{0.32\textwidth}
    \centering
    \includegraphics[width=\textwidth]{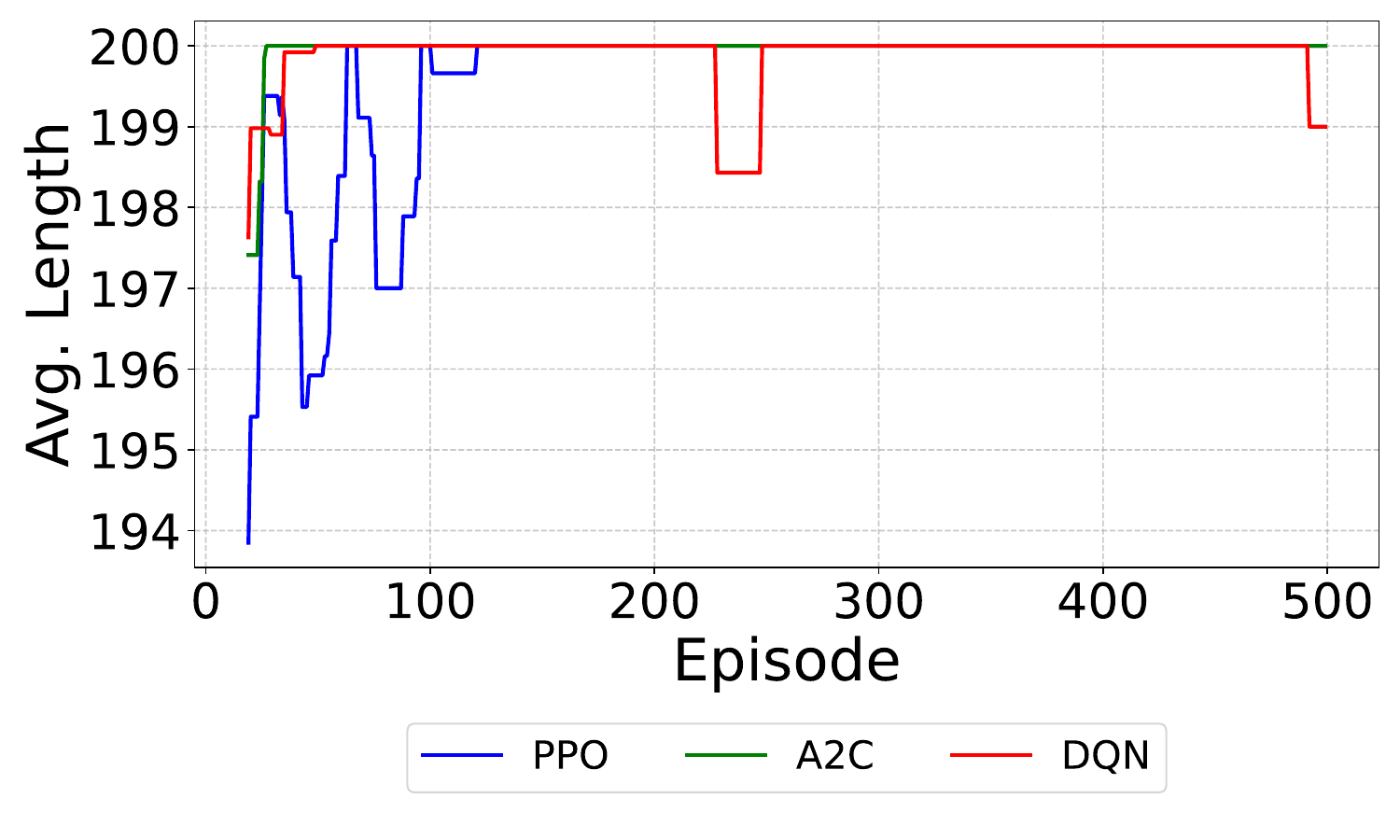}
    \caption{Taxi}
  \end{subfigure}
  \caption{Average episode length during training for SOTA agents (smoothing window = 200).}
  \label{fig: episode length sota training}
\end{figure*}

\paragraph
{RQ1.3: How does the PARL policy compare to state-of-the-art policies?}
We compare our policy to three state-of-the-art methods (PPO, DQN, and A2C; Section~\ref{subsec:related_work_rl}).
We start by comparing the training setups.
Figure~\ref{fig:learning_curves_llm}
presents the average reward during training for the PARL agent and Figure~\ref{fig:learning_curves_rewards_sota} for the SOTA agents.
The two figures show similar patterns for Blackjack and Frozen Lake, where average rewards improve with more training iterations.

During training on the more challenging Taxi task, the SOTA agents learn to minimize negative rewards by avoiding high-penalty actions, such as illegal ``pick-up'' or ``drop-off'' actions, but they still fail to successfully complete the task.
PARL agent initially shows a similar behavior, it then accumulates higher negative rewards per episode, suggesting that adding more examples may introduce noise during training.
We can observe a similar result when comparing the episode length from both the PARL (Figure~\ref{fig:episode length}) and the SOTA policies (Figure~\ref{fig: episode length sota training}).
Notably, a PARL agent learns faster, requiring only a small fraction of the training examples needed by the SOTA agents -- $100$ vs.~$100,000$.

At inference time, Table~\ref{table:Average rewards} shows the average rewards.
In Blackjack, the PARL agent (\textit{Full History}) outperforms the SOTA agents by $16$ pp -- $0.2$ vs.
$0.04$ for PPO, $-0.04$ for DQN, and $-0.02$ for A2C. In Frozen Lake, the performance is similar to the other agents -- $0.66$ vs.~$0.67$ for PPO, $0.61$ for DQN, and $0.62$ for A2C.
However, in the Taxi task, the PARL agent underperforms compared to all SOTA agents.
In addition, none of the agents successfully completed the task, as indicated by the average episode length of 100 -- the maximum number of actions allowed in the environment, resulting in the
truncation/ending of the episode.
The SOTA agents learn to minimize negative rewards by only selecting movement actions ($-1$) and avoiding high-penalty actions.
In contrast, the PARL shows a random behavior,  choosing among all possible actions, including illegal \textit{pick-up} and \textit{drop-off} actions ($-10$).

\subsection
{RQ2: Can a PARL policy benefit from pre-trained knowledge from the LLMs?}
To assess the impact of LLM pre-trained knowledge on PARL, we focus on the \textit{No History} configuration in Table~\ref{table:Average rewards}, where the agent has only access to the current episode’s steps, actions, and rewards. In Blackjack, the agent performs slightly worse than the SOTA models, achieving an average reward of $-0.7$ vs.~$0.04$ (PPO), $-0.04$ (DQN), and $-0.02$ (A2C). This suggests that the LLM’s prior knowledge of card games aids its decision-making process.

However, in Frozen Lake, the absence of context prevents the agent from discovering the correct path, resulting in failure to accomplish the task.
Interestingly, in the Taxi task, the \textit{No History} configuration achieves the highest average reward among all PARL configurations.
This suggests that long context introduces noise, complicating the decision-making process more challenging in complex environments.
The Taxi tasks environment allows up to $100$ steps per episode, compared to an average of only $1.5$
steps per episode in Blackjack (Table~\ref{table:Average rewards}).
This extended episode length may provide a PARL agent with more information to learn the task and to prioritize actions with lower penalties -- similar to the behavior of SOTA agents. In conclusion, a PARL agent benefits from pre-trained knowledge from the LLMs, but this only applies to tasks that are presumably frequent in the training data (e.g., popular card games and video games).

\subsection{RQ3: Does a PARL agent explore and exploit similarly to a standard RL agent?}
\label{sub:RQ3}
Figure~\ref{fig:episode length} examines whether a PARL agent can explore its environment by reporting average episode lengths across three tasks. We focus on Frozen Lake, which requires exploration, unlike Blackjack (where episode length depends on initial cards) and Taxi (where agents reach the maximum actions per episode without solving the task, as discussed in Section~\ref{sub:RQ1}). In Figure~\ref{fig:episode length frozen lake}, we observe an initial increase in episode length from $4$ to $5$, followed by a drop and later stabilization around an average of $7$ steps. A closer analysis reveals that the agent initially follows a repeated action sequence (path), which fails to reach the goal (noted by the drop between episodes $20$ and $30$).
It later discovers a successful path (set of actions) and obtains positive rewards. Occasionally, drops in episode length are due to the task's ``slippery'' nature, where unintended moves may alter the next state (e.g., moving down might result in ending to the left). This analysis shows that the PARL agent is capable of exploration, but it is limited to exploitation.
Once it successfully completes the task, it tends to follow the same set of actions repeatedly, without searching for a more optimal solution.

\section{Conclusion and Future Work}
\label{sec:conclusion}
We introduced PARL, a prompting-based agent that leverages large language models for reinforcement learning in non-linguistic environments. Unlike prior work that focuses on natural language-centered tasks, PARL shows that LLMs can be adapted to basic mathematical decision-making tasks by encoding the environment as textual prompts. Our experiments demonstrate that LLMs can learn from reward-based interactions in simple environments and outperform standard RL baselines in tasks involving common knowledge (e.g., Blackjack). Notably, PARL achieves competitive performance with significantly fewer iterations ($100$ vs.~$100{,}000$), highlighting its sample efficiency. While the PARL exhibits some capacity for exploration, it has limited exploitation behavior. Moreover, it struggles with more complex tasks that require precise computation or abstract state interpretation -- such as decoding grid positions -- underscoring current limitations of LLMs in handling raw numeric values.

This work leads to important future research. One promising direction is the use of retrieval-based methods to select relevant past episodes, as long histories ($h$) may introduce noise to the LLM. In addition,  extending PARL to a multimodal framework incorporating vision or structured graphs may improve its performance in visual tasks (e.g., Atari and card games).

\section{Ethical Considerations}
\label{sec:ethical}
The proposed methodology, Prompt-based Agent for Reinforcement Learning (PARL), introduces an alternative optimization policy by using a large language model through prompting techniques. However, the reliance on LLMs raises several ethical concerns that must be addressed.

Firstly, LLMs are trained on vast and diverse datasets, which can include societal biases or generate inappropriate content. When such biases are integrated into reinforcement learning tasks, they may influence the learning process, potentially leading to harmful or unfair outcomes. This issue is particularly critical when applying LLM-based agents in sensitive or real-world domains, such as healthcare, education, or autonomous systems, where biased or erroneous decisions can have significant consequences.

Secondly, the transparency of LLM decision-making poses a challenge. These models function as black boxes, making it difficult to interpret their reasoning processes and understand how decisions are reached. This lack of interpretability complicates efforts to identify and mitigate biases or errors.

To address these challenges, future research should prioritize the development of techniques to reduce biases in LLMs and improve the interpretability of their decision-making processes. It is important to note that such risks are not inherent to PARL methodology but stem from the base LLMs used.

\section{Limitations}
\label{sec:limitations}

The effectiveness of the proposed PARL method is significantly influenced by
the capabilities of the underlying language models (e.g., GPT, LLaMa-7B-Chat, and Mistral-7B).
While the approach shows potential in using pre-trained LLMs for reinforcement learning tasks, several limitations must be acknowledged.

Firstly, the method struggles with complex tasks that require advanced mathematical reasoning or precise mappings of states and actions. These limitations present a challenge for the model to accomplish the tasks that require the encoding of information.

Secondly, the dependence on text-based prompts introduces scalability challenges. Longer training leads to larger interaction histories, which can result in prompts exceeding the input token limits of LLMs. This may lead to the truncation of important information or limit the number of training episodes. As a result, the agent may struggle to learn in longer tasks or more complicated environments.

Third, environments with large state-action spaces may introduce potential noise. This noise can complicate decision-making and reduce PARL's performance in dynamic or highly variable environments.

Fourth, there is an imbalance in model scale. PARL uses large language models (e.g., GPT-4o), while the RL baselines use much smaller networks. This raises fairness concerns. However, the PARL is trained on only a fraction of the episodes compared to the SOTA baselines, which helps balance the comparison to some extent. Still, we acknowledge that model size can affect performance and encourage future work to explore more size-aligned setups.

Despite these limitations, PARL is the first method to demonstrate an optimization policy using an LLM through prompting techniques. The method has proven effective in generating text-based reinforcement learning strategies, particularly for simpler or knowledge-intensive tasks. However, users should be aware of these limitations when evaluating the method's capabilities and potential applications.

\section*{Acknowledgements}
This work has been supported by a CONACYT
scholarship(2020-000009-01EXTF-00195) and by the project INPROMPT
(Interactive Prompt Optimization with the Human in the Loop for
Natural Language Understanding Model Development and Intervention,
funded by the German Research Foundation, KL 2869/13–1, project
number 521755488).

\section{Bibliographical References}\label{sec:reference}
\bibliographystyle{lrec2026-natbib}
\bibliography{bib}

\appendix
\clearpage
\section{Reinforcement learning environments}\label{appendix:rl_tasks}
The ``Toy text'' environments from the gymnasium
library\footnote{\url{https://gymnasium.farama.org/environments/toy_text}}
are designed to be simple and easy to understand, clearly defined
states and actions. This simplicity makes them ideal for testing and
debugging reinforcement learning algorithms.

\subsection{Blackjack}
The Blackjack environment simulates the classic card game Blackjack. The
player's objective is to accumulate a hand of cards whose total value is as
close to 21 as possible without exceeding it, while simultaneously
outscoring the dealer.

\paragraph{Description.}
The game begins with the dealer holding one card face-up and one face-down,
while the player is dealt two face-up cards. Cards are drawn from an
infinite deck, meaning cards are replaced after each draw. The player's
total is the sum of their cards. They can choose to draw more cards (hit) or
stop drawing (stick). If their total exceeds 21, they go bust and lose
immediately. Once the player sticks, the dealer reveals their face-down card
and continues drawing until their total is at least 17. If the dealer goes
bust, the player wins. If neither the player nor the dealer busts, the
winner is determined by whose total is closer to 21. A tie results in a
draw. Card values are as follows:

\begin{itemize}[noitemsep]
\item Face cards (Jack, Queen, King) are worth 10 points.
\item Aces can count as either $11$ (a ``usable ace'') or 1 point.
\item Numbered cards ($2$--$9$) are worth their face value.
\end{itemize}

\paragraph{Action Space.}
The action space is discrete, consisting of two possible actions:
\begin{itemize}[noitemsep]
\item 0: Stick (end the player's turn).
\item 1: Hit (draw another card).
\end{itemize}

\paragraph{Observation Space.}
The observation space is a tuple of three discrete integers in the format
\texttt{(int, int, int)}:
\begin{itemize}[noitemsep]
\item Player's sum: The total value of the player's cards.
\item Dealer's visible card: The value of the dealer's face-up card.
\item Usable Ace: A binary indicator ($0$ or $1$
  ) showing whether the player has an Ace that can be used as $11$
  without exceeding $21$.
\end{itemize}

\paragraph{Initial State.}
The environment begins with the player's and dealer's card values being
randomly drawn from the deck. The player is handed two cards, while the
dealer receives one face-up card and one face-down card.

\paragraph{Rewards.}
\begin{itemize}[noitemsep]
\item $+1$: The player wins the game.
\item $-1$: The player loses the game.
\item $0$: The game results in a draw.
\end{itemize}

\paragraph{Episode Termination or Truncation.}
The episode ends in any of the following scenarios:
\begin{itemize}[noitemsep]
\item The player's total hand value exceeds 21.
\item The player chooses to stick.
\end{itemize}

\subsection{Frozen Lake}
The Frozen Lake environment simulates a navigation task where the player
must traverse a slippery frozen lake to reach a goal without falling into
holes.

\begin{figure}
  \centering
  \includegraphics[width=\columnwidth]{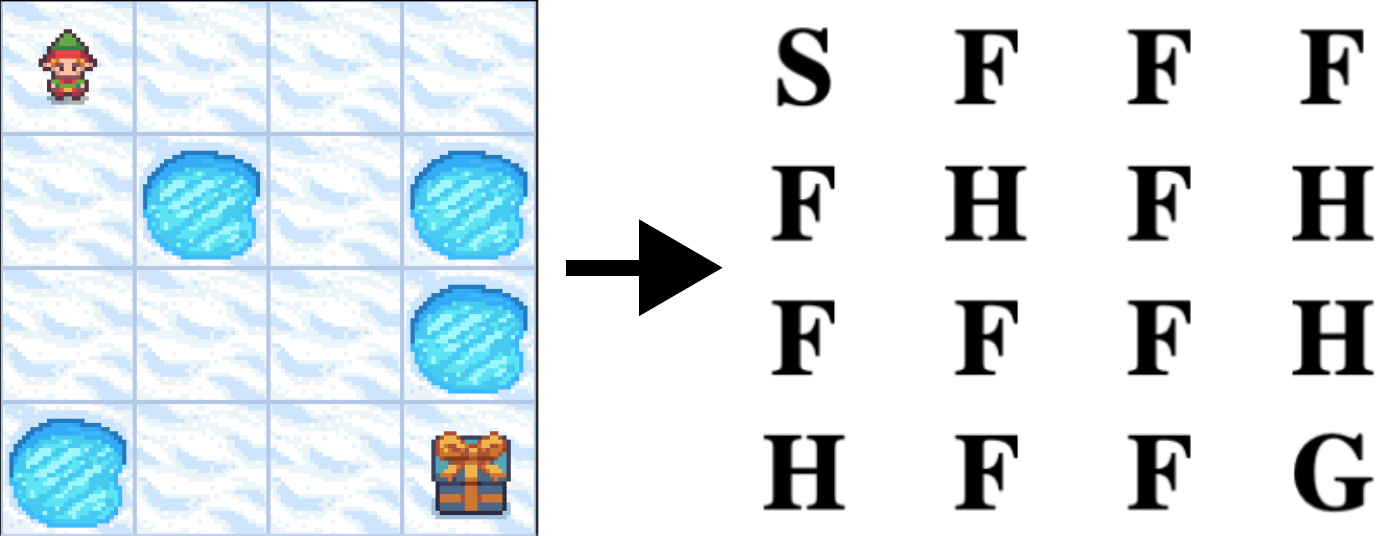}
  \caption
  {Example of a $4 \times 4$ map layout for the FrozenLake environment.}
  \label{fig:FLmap}
\end{figure}

\paragraph{Description.}
The environment consists of a grid representing a frozen lake (Figure~
\ref{fig:FLmap}
shows an example), where each cell corresponds to a specific type of
terrain:
\begin{itemize}[noitemsep]
\item \textbf{S}: Start point (safe).
\item \textbf{F}: Frozen surface (safe but slippery).
\item \textbf{H}: Hole (falling in ends the episode).
\item \textbf{G}: Goal (reaching this ends the episode successfully).
\end{itemize}
Movement is not always deterministic due to the slippery nature of the lake,
which can cause the player to move perpendicular to the intended direction
sometimes. Randomly generated worlds will always have a path to the goal.

\paragraph{Action Space.}
The action space is discrete, with four possible actions:
\begin{itemize}[noitemsep]
\item 0: Move left.
\item 1: Move down.
\item 2: Move right.
\item 3: Move up.
\end{itemize}

\paragraph{Observation Space.}
The observation space is a single discrete integer -- \texttt{int()} -- representing the player's current position on the grid, where both the
row and col start at 0. The position is calculated as:
\begin{equation}
  \text{current\_row} \times \text{ncols} + \text{current\_col}
\end{equation}
For example, in a $4 \times 4$
map, the position of the start cell at the top-left corner can be calculated
as: $0 \times 4 + 0 = 0$.

\paragraph{Initial State.}
The player starts at the \textbf{S} ($0 \times 0$) cell in the grid.

\paragraph{Rewards.}
\begin{itemize}[noitemsep]
\item $+1$: The player reaches the goal (\textbf{G}).
\item $0$: The player falls into a hole (\textbf{H}
  ) or moves across frozen surfaces (\textbf{F}).
\end{itemize}

\paragraph{Episode Termination or Truncation.}
The episode ends in any of the following scenarios:

\begin{itemize}[noitemsep]
\item The player reaches the goal (\textbf{G}).
\item The player falls into a hole (\textbf{H}).
\item The player reaches the maximum movements (e.g., $100$ for
  $4 \times 4$ environment (when using the \texttt{time\_limit} wrapper).
\end{itemize}

\subsection{Taxi}
The Taxi environment simulates a grid-world navigation task where a taxi
must pick up a passenger at one location and drop them at one of four
locations.

\begin{figure}
  \centering
  \includegraphics[width=\columnwidth]{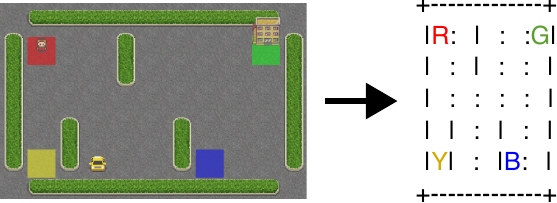}
  \caption{Example the grid map drop-off destination are Red, Blue, Yellow,
    and Green.}
  \label{fig:Taximap}
\end{figure}

\paragraph{Description.}
The environment consists of a $5 \times 5$
grid with specific locations designated by colors Red, Green, Blue, and
Yellow (Figure \ref{fig:Taximap}
show an example of the map). At the start of the game, the taxi, the
passenger, and the destination are randomly initialized. The goal is to
navigate the taxi to the passenger's location, pick them up, and transport
them to the destination.
Movement in the grid is deterministic, with walls that act as obstacles
between certain cells. The agent must navigate around these walls to
complete the task. Actions such as attempting to pick up or drop off a
passenger in the wrong location result in penalties.

\paragraph{Action Space.}
The action space is discrete, consisting of six possible actions:
\begin{itemize}[noitemsep]
\item 0: Move south (down).
\item 1: Move north (up).
\item 2: Move east (right).
\item 3: Move west (left).
\item 4: Pick up the passenger.
\item 5: Drop off the passenger.
\end{itemize}

\paragraph{Observation Space.}
The environment contains 500 discrete states, determined by $25$
possible taxi positions, $5$
possible passenger locations (including the scenario where the passenger is
already in the taxi), and $4$
destination points. Destinations on the map are identified by the first
letter of their corresponding color. Passenger Locations:
\begin{itemize}[noitemsep]
\item 0: Red
\item 1: Green
\item 2: Yellow
\item 3: Blue
\item 4: In the taxi
\end{itemize}

Destinations:
\begin{itemize}[noitemsep]
\item 0: Red
\item 1: Green
\item 2: Yellow
\item 3: Blue
\end{itemize}

An observation is represented as an integer that encodes the state using the
formula:
\begin{equation}
  \begin{split}
    ((\text{taxi\_row} \times 5 + \text{taxi\_col}) \times 5 + \\
    \text{passenger\_location})
    \times 4 + \text{destination}
  \end{split}
\end{equation}
There are $400$
states that can be accessed during an episode. The missing states occur when
the passenger's location matches their destination, which typically marks
the end of the episode. Additionally, four extra states are observed
immediately after a successful episode, where both the taxi and the
passenger are at the destination. This results in a total of $404$
discrete reachable states.

\paragraph{Initial State.}
The taxi's initial position, the passenger's location, and the destination
are all randomly assigned at the start of the episode.

\paragraph{Rewards.}
\begin{itemize}[noitemsep]
\item $+20$: Successfully dropping off the passenger at the destination.
\item $-10$: Attempting an illegal pickup or drop-off.
\item $-1$: Each time step to encourage efficiency.
\end{itemize}

\paragraph{Episode Termination or Truncation.}
The episode ends in either of the following scenarios:
\begin{itemize}[noitemsep]
\item The passenger is successfully dropped off at the destination.
\item The maximum number of steps is reached, truncating the episode.
\end{itemize}

\section{PARL: Prompt-based Agent Optimization for Reinforcement Learning}
\label{appendix: PromptRL-Agent}

PARL introduces a novel reinforcement learning method leveraging large
language models (LLMs) as decision-making agents through prompting
techniques. The method integrates the task description $\mathcal{T}$
with the agent's interaction history ($h_t$
) with the environment to define the policy. The PARL policy is defined as:
\[
  \mathcal{P}^{\text{PARL}} = \mathcal{T} \bigoplus_{t=0}^n h_t,
\]
where $\mathcal{T}$ represents the task description and $h_t$
denotes the interaction history.

The combination of $\mathcal{T}$ and $h_t$
enables LLMs to make decisions based on interactions with the environment.
Examples of PARL agent prompts are presented in Section~
\ref{appendix:blackjack prompt} for Blackjack, Section~
\ref{appendix:frozen lake prompt} for Frozen Lake, and Section~
\ref{appendix:taxi prompt}
for Taxi. Detailed explanations for each task can be found in their
respective sections.

\subsection{BlackJack Prompt}
\label{appendix:blackjack prompt}
\textbf{Task description:}
\begin{lstlisting}
Imagine you are a RL agent for a black Jack Agent. The task is the following.

Description
The game starts with the dealer having one face up and one face down card, wh
ile the player has two face up cards. All cards are drawn from an infinite de
ck (i.e. with replacement).
The card values are:
* Face cards (Jack, Queen, King) have a point value of 10.
* Aces can either count as 11 (called a 'usable ace') or 1.
* Numerical cards (2-9) have a value equal to their number.
The player has the sum of cards held. The player can request additional cards
(hit) until they decide to stop (stick) or exceed 21 (bust, immediate loss).
After the player sticks, the dealer reveals their facedown card, and draws ca
rds until their sum is 17 or greater. If the dealer goes bust, the player wins.
If neither the player nor the dealer busts, the outcome (win, lose, draw) is
decided by whose sum is closer to 21.

Action Space
The action shape is (1,) in the range {0, 1} indicating whether to stick or hit.
* 0: Stick
* 1: Hit

Observation Space
The observation consists of a 3-tuple containing: the player's current sum, t
he value of the dealer's one showing card (1-10 where 1 is ace), and whether
the player holds a usable ace (0 or 1).
The observation is returned as (int(), int(), int()).

Rewards
* win game: +1
* lose game: -1
* draw game: 0
* win game with natural blackjack: +1.5 (if natural is True) +1 (if natural i
s False)

Episode End
The episode ends if the following happens:
* Termination:
1. The player hits and the sum of hand exceeds 21.
2. The player sticks.
An ace will always be counted as usable (11) unless it busts the player.

Example of previous episodes:
\end{lstlisting}
\textbf{History:}
\begin{lstlisting}
--- Episode 0 --
Previous observation within the episode:
---Step: 0---
observations: (20, 10, 0)
action taken: 0
Result:
observations: (20, 10, 0)
reward: 1.0
terminated: True
truncated: False
Episode 0 end: Episode reward 1.0

--- Episode 1 --
Previous observation within the episode:
---Step: 0---
observations: (19, 5, 0)
action taken: 0
Result:
observations: (19, 5, 0)
reward: -1.0
terminated: True
truncated: False
Episode 1 end: Episode reward -1.0

--- Episode 2 --
Previous observation within the episode:
---Step: 0---
observations: (7, 9, 0)
action taken: 1
Result:
observations: (14, 9, 0)
reward: 0.0
terminated: False
truncated: False
---Step: 1---
observations: (14, 9, 0)
action taken: 1
Result:
observations: (18, 9, 0)
reward: 0.0
terminated: False
truncated: False
---Step: 2---
observations: (18, 9, 0)
action taken: 0
Result:
observations: (18, 9, 0)
reward: 1.0
terminated: True
truncated: False
Episode 2 end: Episode reward 1.0

--- Episode 3 --
Previous observation within the episode:
---Step: 0---
observations: (18, 10, 0)
action taken: 0
Result:
observations: (18, 10, 0)
reward: 1.0
terminated: True
truncated: False
Episode 3 end: Episode reward 1.0

--- Episode 4 --
Previous observation within the episode:
---Step: 0---
observations: (18, 9, 0)
action taken: 0
Result:
observations: (18, 9, 0)
reward: -1.0
terminated: True
truncated: False
Episode 4 end: Episode reward -1.0

--- Episode 5 --
Previous observation within the episode:
---Step: 0---
observations: (7, 6, 0)
action taken: 1
Result:
observations: (17, 6, 0)
reward: 0.0
terminated: False
truncated: False
---Step: 1---
observations: (17, 6, 0)
action taken: 0
Result:
observations: (17, 6, 0)
reward: 1.0
terminated: True
truncated: False
Episode 5 end: Episode reward 1.0
\end{lstlisting}

\subsection{Frozen Lake Prompt}
\label{appendix:frozen lake prompt}

\textbf{Task description:}
\begin{lstlisting}
Imagine you are a Reinforcement Learning Agent for the  following task.

Frozen lake involves crossing a frozen lake from start to goal without fallin
g into any holes by walking over the frozen lake. The player may not always m
ove in the intended direction due to the slippery nature of the frozen lake.

Description
The game starts with the player at location [0,0] of the frozen lake grid wor
ld with the goal located at far extent of the world e.g. [3,3] for the 4x4 en
vironment.
Holes in the ice are distributed in set locations when using a pre-determined
map or in random locations when a random map is generated.
The player makes moves until they reach the goal or fall in a hole.
The lake is slippery (unless disabled) so the player may move perpendicular t
o the intended direction sometimes (see is_slippery).
Randomly generated worlds will always have a path to the goal.

Action Space
The action shape is (1,) in the range {0, 3} indicating which direction to mo
ve the player.
0: Move left
1: Move down
2: Move right
3: Move up

Observation Space
The observation is a value representing the player's current position as curr
ent_row * nrows + current_col (where both the row and col start at 0).
For example, the goal position in the 4x4 map can be calculated as follows: 3
* 4 + 3 = 15. The number of possible observations is dependent on the size of
the map.
The observation is returned as an int().

Starting State
The episode starts with the player in state [0] (location [0, 0]).

Rewards
Reach goal: +1
Reach hole: 0
Reach frozen: 0

Episode End
The episode ends if the following happens:

Termination:
The player moves into a hole.
The player reaches the goal at max(nrow) * max(ncol) - 1 (location [max(nrow)
-1, max(ncol)-1]).
Truncation (when using the time_limit wrapper):
The length of the episode is 100 for 4x4 environment, 200 for FrozenLake8x8-v
1 environment.

Example of previous episodes:
\end{lstlisting}

\textbf{History:}
\begin{lstlisting}
--- Episode 0 --
Previous position within the episode:
---Step: 0---
Position: 0
action taken: right
Result:
Position: 1
reward: 0.0
terminated: False
truncated: False
---Step: 1---
Position: 1
action taken: right
Result:
Position: 2
reward: 0.0
terminated: False
truncated: False
---Step: 2---
Position: 2
action taken: right
Result:
Position: 3
reward: 0.0
terminated: False
truncated: False
---Step: 3---
Position: 3
action taken: down
Result:
Position: 7
reward: 0.0
terminated: True
truncated: False
Episode 0 end: Episode reward 0.0

--- Episode 1 --
Previous position within the episode:
---Step: 0---
Position: 0
action taken: right
Result:
Position: 1
reward: 0.0
terminated: False
truncated: False
---Step: 1---
Position: 1
action taken: right
Result:
Position: 2
reward: 0.0
terminated: False
truncated: False
---Step: 2---
Position: 2
action taken: right
Result:
Position: 3
reward: 0.0
terminated: False
truncated: False
---Step: 3---
Position: 3
action taken: down
Result:
Position: 7
reward: 0.0
terminated: True
truncated: False
Episode 1 end: Episode reward 0.0

--- Episode 2 --
Previous position within the episode:
---Step: 0---
Position: 0
action taken: right
Result:
Position: 1
reward: 0.0
terminated: False
truncated: False
---Step: 1---
Position: 1
action taken: right
Result:
Position: 2
reward: 0.0
terminated: False
truncated: False
---Step: 2---
Position: 2
action taken: right
Result:
Position: 3
reward: 0.0
terminated: False
truncated: False
---Step: 3---
Position: 3
action taken: down
Result:
Position: 7
reward: 0.0
terminated: True
truncated: False
Episode 2 end: Episode reward 0.0
\end{lstlisting}

\subsection{Taxi Prompt}
\label{appendix:taxi prompt}

\textbf{Task description:}
\begin{lstlisting}
Imagine you are a Reinforcement Learning Agent for the  following task:

The Taxi Problem involves navigating to passengers in a grid world, picking them up and dropping them off at one of four locations.

Description
There are four designated pick-up and drop-off locations (Red, Green, Yellow
and Blue) in the 5x5 grid world. The taxi starts off at a random square and the passenger at one of the designated locations.
The goal is move the taxi to the passenger's location, pick up the passenger,
move to the passenger's desired destination, and drop off the passenger. Once
the passenger is dropped off, the episode ends.
The player receives positive rewards for successfully dropping-off the passen
ger at the correct location. Negative rewards for incorrect attempts to pick-up/drop-off passenger and for each step where another reward is not received.

Observation Space
There are 500 discrete states since there are 25 taxi positions, 5 possible locations of the passenger (including the case when the passenger is in the taxi), and 4 destination locations.
Destination on the map are represented with the first letter of the color.

Passenger locations:
0: Red
1: Green
2: Yellow
3: Blue
4: In taxi

Destinations:
0: Red
1: Green
2: Yellow
3: Blue

An observation is returned as an int() that encodes the corresponding state,
calculated by ((taxi_row * 5 + taxi_col) * 5 + passenger_location) * 4 + dest
ination Note that there are 400 states that can actually be reached during an episode. The missing states correspond to situations in which the passenger is at the same location as their destination, as this typically signals the end of an episode. Four additional states can be observed right after a successful episodes, when both the passenger and the taxi are at the destination. This gives a total of 404 reachable discrete states.

Starting State
The episode starts with the player in a random state.

Rewards
-1 per step unless other reward is triggered.
+20 delivering passenger.
-10 executing 'pickup' and 'drop-off' actions illegally.
An action that results a noop, like moving into a wall, will incur the time step penalty. Noops can be avoided by sampling the action_mask returned in info.

Episode End
The episode ends if the following happens:
Termination: 1. The taxi drops off the passenger.
Truncation (when using the time_limit wrapper): 1. The length of the episode
is 200.

Map:
    +---------+
    |R: | : :G|
    | : | : : |
    | : : : : |
    | | : | : |
    |Y| : |B: |
    +---------+

Action Space
The action shape is (1,) in the range {0, 5} indicating which direction to move the taxi or to pickup/drop off passengers.
0: Move south (down)
1: Move north (up)
2: Move east (right)
3: Move west (left)
4: Pickup passenger (pickup)
5: Drop off passenger (drop_off)

Example of previous episodes:
\end{lstlisting}

\textbf{History:}
\begin{lstlisting}
--- Episode 0 --
Previous position within the episode:
---Step: 0---
Observation: 201
action taken: right
Result:
Observation: 221
reward: -1
terminated: False
truncated: False
Episode accumulative reward -1
---Step: 1---
Observation: 221
action taken: pickup
Result:
Observation: 221
reward: -10
terminated: False
truncated: False
Episode accumulative reward -11
---Step: 2---
Observation: 221
action taken: pickup
Result:
Observation: 221
reward: -10
terminated: False
truncated: False
Episode accumulative reward -21
---Step: 3---
Observation: 221
action taken: down
Result:
Observation: 321
reward: -1
terminated: False
truncated: False
Episode accumulative reward -22
---Step: 4---
Observation: 321
action taken: right
Result:
Observation: 341
reward: -1
terminated: False
truncated: False
Episode accumulative reward -23
---Step: 5---
Observation: 341
action taken: right
Result:
Observation: 341
reward: -1
terminated: False
truncated: False
Episode accumulative reward -24
---Step: 6---
Observation: 341
action taken: left
Result:
Observation: 321
reward: -1
terminated: False
truncated: False
Episode accumulative reward -25
---Step: 7---
Observation: 321
action taken: pickup
Result:
Observation: 321
reward: -10
terminated: False
truncated: False
Episode accumulative reward -35
---Step: 8---
Observation: 321
action taken: right
Result:
Observation: 341
reward: -1
terminated: False
truncated: False
Episode accumulative reward -36
---Step: 9---
Observation: 341
action taken: left
Result:
Observation: 321
reward: -1
terminated: False
truncated: False
Episode accumulative reward -37
---Step: 10---
Observation: 321
action taken: pickup
Result:
Observation: 321
reward: -10
terminated: False
truncated: False
Episode accumulative reward -47
\end{lstlisting}

\end{document}